\documentclass{article}

\usepackage{iclr2027_conference,times}
\usepackage[T1]{fontenc}
\iclrfinalcopy
\usepackage{amsmath,amssymb,amsfonts}
\usepackage{booktabs,array,multirow,graphicx}
\usepackage{url,hyperref}

\title{HeadGuard: Selective Head Protection\\ for Low-Bit VLM KV-Cache Quantization}
\author{Nenad Banfic\\CoreAI, Microsoft}

\begin{document}
\maketitle

\begin{abstract}
Low-bit key--value (KV) cache quantization saves storage but can sharply degrade
vision--language model (VLM) accuracy. We introduce \textbf{HeadGuard}, a
composable head-protection method that augments a base KV-cache quantizer
with a fixed high-precision mask. Image-sensitivity and output-sensitivity scores select
physical KV heads offline, with approximately $1/8$ protected in the main experiments;
their image keys and optionally values remain in bfloat16 (BF16), while the base quantizes
unprotected image entries.
Across eight VLMs, three base quantizers, and eight benchmarks (six
discriminative and two generative), HeadGuard recovers a substantial fraction of
lost accuracy on weaker quantizers, with the strongest gains for Qwen and
InternVL. At $2$ bits, the six-task discriminative mean over eight models rises from $0.436$ to
$0.580$ on the weakest base; protection can also improve generated answers and caption fidelity to BF16 outputs. Mean accuracy gains persist across all three quantizers with both
tested calibration datasets. Keys-only protection retains substantial recovery
at lower modeled storage cost. Evaluated through simulated quantization, HeadGuard offers a
composable way to improve low-bit VLM accuracy without replacing the underlying
quantizer.
\end{abstract}

\section{Introduction}
Image tokens often dominate a vision--language model's KV cache. Low-bit
quantization reduces storage~\citep{kivi,kvquant}, but its effect on accuracy
varies sharply. Per-token min--max, our simplest base quantizer, sets each
token's quantization range from its own minimum and maximum and rounds every
entry to the nearest level in that range (round-to-nearest): on POPE, per-token $2$-bit min--max gives $0.832$ accuracy for
LLaVA-NeXT and $0.496$ for Qwen2.5-VL under the same evaluation protocol.

Preserving a few entries in high precision is a common remedy. We ask when
protecting image KV heads helps, compared with changing the base quantizer or
increasing precision. Throughout, we quantize only the KV cache and keep model
weights in bfloat16 (BF16). Our BF16 reference leaves the entire KV cache in BF16 as well,
avoiding the rounding error that low-bit cache quantization introduces.

We introduce \textbf{HeadGuard}, a composable protection method, not a replacement
quantizer. It ranks heads offline using image-sensitivity and output-sensitivity
scores, which measure how predictions respond to rescaling a head's image
contribution and its entire output, respectively. Selected image keys---and
optionally values---remain in BF16, while the base quantizer handles the remaining
image entries. The resulting mask is
reused across tasks without weight updates or inference-time gradients.

Across eight VLMs, three base quantizers, and eight benchmarks, we examine when
head protection helps and how its gains depend on precision, head budget, and
calibration. We also test whether image sensitivity adds value beyond output
sensitivity and whether protecting keys alone retains recovery.

\section{Related Work}
\textbf{KV-cache quantization.} KIVI~\citep{kivi} quantizes keys per channel
and values per token; KVQuant~\citep{kvquant} combines low-bit cache quantization
with outlier handling. AKVQ-VL~\citep{akvq} uses attention-based token saliency
to allocate precision and Hadamard transforms to reduce outliers. Our KIVI-style
and AKVQ-style bases implement selected arithmetic components in a shared
harness, not the complete published algorithms or serving kernels.

\textbf{Sensitivity-based precision allocation.} Head pruning can use
loss sensitivity to head masks~\citep{sixteenheads}. KVmix~\citep{kvmix} uses
gradients of key/value projection weights to choose bit-widths for each layer.
RateQuant~\citep{ratequant} chooses separate bit-widths for each head's keys and
values under a total bit budget. It uses squared gradients of KV projection
outputs and fitted, quantizer-specific curves that estimate quantization error
at different bit-widths. Both target text LLMs.
HeadGuard also uses sensitivity to guide precision allocation, but targets image
KV in VLMs. It combines image-sensitivity and output-sensitivity scores to select
a fixed set of top-$K$ KV heads, preserving their image keys and optionally values
in BF16. Unlike RateQuant, it does not fit distortion curves; text KV remains BF16.

\textbf{Head-based token retention.} HeadKV~\citep{headkv} allocates token budgets
from retrieval and reasoning importance; DuoAttention~\citep{duoattention} keeps
full context for retrieval heads and a short cache for streaming heads; and
SparseMM~\citep{sparsemm} uses OCR-grounded attention scores for per-head token
budgets. Our attention-mass and visual-response controls are simplified proxies,
not AKVQ-VL's or SparseMM's original scoring rules.
They let us compare head-ranking signals using the same calibration data and
BF16 protection budget.

\textbf{Visual token reduction.} DART~\citep{dart} removes visual tokens that
carry similar information to other tokens. Its experiments show that choosing
which tokens to keep using attention scores can perform worse than choosing
them at random.
Unlike such pruning, HeadGuard keeps all tokens and only changes their KV
precision.

\section{HeadGuard}
\label{sec:method}
HeadGuard operates on image-token KV entries in the VLM's language-model attention
layers.
HeadGuard takes a frozen VLM, a labeled calibration set $\mathcal D$, a head
budget $K$, and a base quantizer $\mathcal Q$. It scores heads offline, selects
a fixed set, and bypasses low-bit quantization for selected entries.

\textbf{Two sensitivity scores.} During calibration, we attach two differentiable
scalar gates to each query head $(l,h)$, both evaluated at $1$. The image gate
$\gamma_{l,h}$ scales the post-softmax attention weights in columns corresponding
to known image-token positions in the input sequence, across all query positions
in that head and \emph{without renormalization}. This scaling occurs before the
attention weights are multiplied by the value matrix. The output gate
$\eta_{l,h}$ instead scales the head's entire output after this weighted sum and
before the shared attention output projection.
Writing a head output as $u=u_{\rm img}+u_{\rm text}$, with contributions from image
and text positions respectively, both gates are present in the same forward pass
and give
\[
u(\gamma,\eta)=\eta(\gamma u_{\rm img}+u_{\rm text}).
\]
With $\eta=1$, the image gate rescales only the image contribution without
reallocating attention to text. With $\gamma=1$, the output gate scales the whole
head output.
Unlike the output gate, the image gate leaves the text contribution unscaled.
We measure local sensitivity to these perturbations using mean absolute
gradients~\citep{sixteenheads}:
\begin{equation}
g_{l,h}=\mathbb E_{x\sim\mathcal D}\left|\frac{\partial\,\log P(a^\star\mid x)}{\partial\gamma_{l,h}}\right|,
\qquad
o_{l,h}=\mathbb E_{x\sim\mathcal D}\left|\frac{\partial\,\mathcal L_{\rm prompt}(x)}{\partial\eta_{l,h}}\right|.
\label{eq:scores}
\end{equation}
Each expectation $\mathbb E_{x\sim\mathcal D}$ averages the per-example absolute
gradient over the calibration set $\mathcal D$.
The image-sensitivity score $g_{l,h}$ measures the mean absolute sensitivity of the
correct-answer log-probability $\log P(a^\star\mid x)$ to the image gate.
The output-sensitivity score $o_{l,h}$ measures the corresponding sensitivity of
$\mathcal L_{\rm prompt}$, the mean next-token cross-entropy on the text token
positions, to the output gate; the image tokens stay in the input and are only
excluded as prediction targets.
We pair these objectives to assess answer sensitivity to image contributions
alongside whole-head sensitivity across prompt-text predictions. The selector
comparison does not isolate gate choice from objective choice.
Computing both scores for all heads uses one forward pass and
two backward passes through the same graph per calibration example: one computes
the answer-log-probability gradients with respect to the image gates, and the
other computes the prompt-loss gradients with respect to the output gates.
Neither gates nor model weights are updated. Both are local (first-order) sensitivity
signals we use only to rank heads, not measures of faithful visual reasoning or of
actual quantization error.

\textbf{Selecting physical KV heads.} Grouped-query attention shares one KV
head among several query heads~\citep{gqa}, so we average the query-head scores
in each physical KV group to obtain image-sensitivity and output-sensitivity scores
$\bar g_i$ and $\bar o_i$. Here $i$ indexes physical KV heads across all
language-model attention layers, and $N_{\rm KV}$ is their total count.
The two scores live on different scales, so we standardize each to
zero mean and unit variance across the model's KV heads ($z(\cdot)$), add them,
and protect the $K$ heads with the largest sum:
\begin{equation}
s_i=z(\bar g_i)+z(\bar o_i),\qquad
\mathcal P=\operatorname{TopK}(s,K).
\label{eq:selection}
\end{equation}
The head budget $K$ is a parameter: our main K+V and K-only experiments use
$K=\operatorname{round}(N_{\rm KV}/8)$, and the capped budget sweep varies the
fraction. We choose this fraction to keep most image KV entries quantized while
reserving a limited BF16 budget for sensitive heads.

\textbf{Applying protection.} With key-and-value (K+V) protection, selected heads
keep both image keys and values in BF16. With keys-only (K-only) protection,
selected image keys stay in BF16, while image values follow the base quantizer.
Text keys and values remain in BF16 in both cases.
A compatible base must allow quantization to be skipped for individual KV heads.

\section{Evaluation Scope and Protocol}
\label{sec:protocol}
\textbf{Models and tasks.} We evaluate LLaVA-1.5-7B~\citep{llava,llava15},
LLaVA-NeXT-Mistral-7B~\citep{llavanext}, Qwen2.5-VL-7B~\citep{qwen25vl},
Qwen2-VL-2B~\citep{qwen2vl}, Gemma-3-12B~\citep{gemma3}, Gemma-4-12B~\citep{gemma4},
Pixtral-12B~\citep{pixtral}, and InternVL3-8B~\citep{internvl3}. The six discriminative
tasks are POPE~\citep{pope}, MME~\citep{mme}, MMBench~\citep{mmbench},
image-based ScienceQA~\citep{scienceqa}, multiple-choice MMMU~\citep{mmmu}, and
SEED-Bench~\citep{seedbench}. TextVQA~\citep{textvqa} and COCO
captioning~\citep{coco} add free-form generation. Unless noted otherwise, head
scores are computed for each model and evaluation protocol from $100$ labeled
POPE examples and then frozen across tasks and base quantizers. The calibration
comparison in \S\ref{sec:ablation} additionally computes scores from $100$ labeled
MMBench examples.

\textbf{Evaluation protocols.}
A preliminary check on Qwen2-VL-2B motivated the keys-only evaluation:
quantizing only the image keys collapsed accuracy, whereas quantizing only the
image values left it near BF16.
\begin{enumerate}
\item \emph{K+V protection:} $2$--$3$ bits, a fixed $1/8$ head budget,
all eligible discriminative examples after holdouts.
\item \emph{K-only evaluation:} $2$ bits, the same head budget and evaluation
scope, with only selected keys protected.
\item \emph{Budget/precision ablation:} $2$--$4$-bit K+V protection at nominal fractions
$1/16$, $1/12$, $1/8$, and $1/4$, plus uniform, on capped discriminative subsets,
also comparing POPE and MMBench calibration.
\end{enumerate}
The first two protocols give our main results; the third is a supporting
ablation (\S\ref{sec:ablation}).
For generation, the main K+V and K-only evaluations use $1000$ TextVQA examples
and $200$ captioning examples; the capped sweeps use $200$ examples for each of
these two tasks.

\textbf{Quantization setup.} We test per-token min--max for keys and values;
KIVI-style, with per-channel keys and per-token values; and AKVQ-style, with
query/key rotation and per-token quantization.
In each KV head, AKVQ-style additionally keeps the keys and values of a
nominal $2\%$ of image tokens in BF16, selecting those with the largest key norms.
This applies with or without HeadGuard.
KIVI-style omits the original grouped/residual-cache policy and recomputes key
ranges over the supplied sequence, including text; text KV nevertheless remains
in BF16.
Our large sweeps simulate low-bit quantization inside attention: entries assigned
low precision are rounded to the base quantizer's levels and reconstructed as
floating-point values, rather than stored in a packed low-bit cache. These
experiments measure the effect on accuracy, not the memory savings or speed of a
packed cache.

\textbf{Head-selection baselines.} To assess the contribution of the hybrid score
$s$, we compare against selectors that protect the same number $K$ of physical
KV heads. The score-based baselines use image sensitivity $g$
alone; output sensitivity $o$ alone; activation magnitude (the squared head-output
norm); image-attention mass (the sum of post-softmax attention weights on image
keys, an AKVQ-inspired proxy~\citep{akvq}); or visual-response norm (the norm of
the head output contributed by image tokens, a proxy inspired by SparseMM's
visual-head analysis~\citep{sparsemm}). For the three non-gradient signals, we
average over query-token positions and aggregate over calibration examples.
As in HeadGuard, all score-based baselines then average scores across query heads
sharing a physical KV head before selecting the top-$K$ KV heads.
We also compare against random head selection on discriminative tasks.
Implementation details for all selectors, including score aggregation and head
selection, are provided in the code supplement.

\textbf{Data splits, metrics, and averaging.} On POPE, head scores are calibrated
once, on $100$ questions drawn in a fixed shuffled order. For every base,
main K+V and K-only results exclude those calibration questions and the same
$8$ questions reserved for preliminary diagnostic checks, leaving $8892$.
Calibration and evaluation questions are disjoint.

After filtering and holdouts, the
numbers of examples used for main discriminative evaluation are: POPE $8892$,
MME $2366$, MMBench $4321$,
ScienceQA $2009$, MMMU $839$, and SEED $14{,}225$. The capped
budget sweep uses up to $1000$ evaluation questions per discriminative task and
$200$ examples per generative task.
In the capped sweep, we evaluate on $900$ held-out POPE questions when
calibrating on POPE, and $900$ held-out MMBench questions when calibrating on
MMBench. These evaluation sets exclude the $100$ questions used for calibration.
Within each setting, the held-out set is identical across the three quantizer
bases.
For the discriminative tasks we score each example from the model's logits on the
candidate answer tokens (yes/no or the option letters); the same scoring rule is
used across models and bases. TextVQA
uses a VQA-style match to the human answer annotations
with case, punctuation, and article normalization and a token-containment
fallback (not the official VQA scorer). Captioning uses case-sensitive whitespace-token
ROUGE-L F1~\citep{rouge} against BF16 output, not human captions, with no
lowercasing or punctuation removal; BF16 scores $1$
by convention. We never average these generative scores with classification
accuracy. Greedy decoding generates at most $12$ new tokens for TextVQA and $64$
for captions, excluding input tokens and stopping earlier at an end-of-sequence
token.

For each model, we average accuracy across the six discriminative tasks; the
overall mean then gives each model equal weight. These averages use task scores
rounded to three decimal places, with final rounding to the nearest reported
digit (ties rounded up).
Overall means and gains use the aggregates before this final rounding.

\textbf{Preprocessing.} We shrink each image so its longest side is at most
$336$ pixels for LLaVA and $768$ for Qwen, Pixtral, and InternVL, turn off
Gemma's image-cropping mode (pan-and-scan), and limit InternVL to one image tile;
each image therefore yields a bounded number of tokens. Gemma mixes short-range and long-range
attention, so we cap the image length and do not test very long image inputs.

\section{When Does Head Protection Help?}
\subsection{Recovery depends on the model and precision}
Table~\ref{tab:main} reports K+V protection on the per-token min--max base,
averaged over all six discriminative tasks for all eight models. At $2$ bits,
HeadGuard lifts the eight-model mean from $0.436$ to $0.580$---a $14.4$-point
gain. Qwen2.5-VL rises from $0.360$ to $0.747$ and InternVL3 from $0.362$
to $0.688$, while Pixtral recovers only part of its loss (from $0.538$ to $0.642$;
BF16 $0.770$). Neither Gemma model recovers from $2$-bit min--max degradation
under this mask. At $3$ bits, the unprotected mean is $0.564$, but Qwen and
InternVL remain fragile; protection raises the overall mean to $0.724$.
Gemma-3 also improves (from $0.651$ to $0.693$), whereas LLaVA and Pixtral change little.

\begin{table}[t]
\centering
\caption{\textbf{Head protection recovers low-bit min--max accuracy, unevenly.}
HeadGuard protects selected image keys and values on the per-token min--max base, averaged over
POPE/MME/MMBench/ScienceQA/MMMU/SEED.
$K\approx N_{\rm KV}/8$. BF16 is a reference, not an upper bound.}
\label{tab:main}
\setlength{\tabcolsep}{6pt}
\begin{tabular}{@{}lccccc@{}}
\toprule
 & \multicolumn{2}{c}{$2$-bit} & \multicolumn{2}{c}{$3$-bit} & \\
\cmidrule(lr){2-3}\cmidrule(lr){4-5}
Model & Uniform & HeadGuard & Uniform & HeadGuard & BF16 \\
\midrule
LLaVA-1.5 & 0.513 & 0.579 & 0.641 & 0.642 & 0.643 \\
LLaVA-NeXT & 0.621 & 0.657 & 0.667 & 0.666 & 0.665 \\
Qwen2.5-VL & 0.360 & 0.747 & 0.361 & 0.781 & 0.790 \\
Qwen2-VL & 0.363 & 0.598 & 0.359 & 0.729 & 0.731 \\
Gemma-3 & 0.372 & 0.367 & 0.651 & 0.693 & 0.754 \\
Gemma-4 & 0.362 & 0.363 & 0.705 & 0.712 & 0.786 \\
Pixtral & 0.538 & 0.642 & 0.764 & 0.766 & 0.770 \\
InternVL3 & 0.362 & 0.688 & 0.367 & 0.802 & 0.826 \\
\bottomrule
\end{tabular}
\end{table}

\subsection{Base quality matters more than recovery alone}
Table~\ref{tab:fullkv_selectors} compares all three bases on the same set of
model--task cells. At $2$ bits, HeadGuard gains $14.37$ points on min--max,
$12.73$ on AKVQ-style, and $0.85$ on KIVI-style. Average gains are larger
on the weaker bases. KIVI-style is robust on most models, leaving limited
headroom for further gains from protection. A large gain does not guarantee the best
absolute accuracy: protected min--max trails \emph{unprotected} KIVI-style by
$10.67$ points at $2$ bits, so a repaired weak base can still lose to a strong
one. Head protection is therefore a targeted repair, not a substitute for
choosing a strong base.

\begin{table}[t]
\centering
\caption{\textbf{K+V protection: base accuracy and six-task selector gains.}
Uniform gives each base's absolute six-task accuracy; the remaining columns are
percentage-point gains over that uniform arm, $K\approx N_{\rm KV}/8$.
Each row averages the six tasks and eight models equally, using full eligible
task evaluation. Image and Output use only the image-sensitivity score $g$
and output-sensitivity score $o$,
respectively; HeadGuard combines them as $s=z(\bar g)+z(\bar o)$.
Magn., Attn., and Vis. use activation magnitude, image-attention mass, and
visual-response norm. Random head selection is reported for the min--max base
only. These are descriptive means,
not significance tests; small differences can reflect rounding.}
\label{tab:fullkv_selectors}
\setlength{\tabcolsep}{4pt}
\begin{tabular}{@{}clrrrrrrrr@{}}
\toprule
 & & \multicolumn{1}{c}{Accuracy} & \multicolumn{7}{c}{Gain (percentage points)} \\
\cmidrule(lr){3-3}\cmidrule(lr){4-10}
Bits & Base & Uniform & Image & Output & Magn. & Attn. & Vis. & Random & HeadGuard \\
\midrule
2 & min--max & 0.436 & 2.57 & 13.34 & 1.26 & 6.64 & 1.34 & 1.39 & 14.37 \\
2 & KIVI-style & 0.687 & 1.02 & 0.60 & 0.35 & 0.61 & 0.48 & -- & 0.85 \\
2 & AKVQ-style & 0.601 & 3.18 & 12.74 & 0.32 & 4.51 & 0.50 & -- & 12.73 \\
\midrule
3 & min--max & 0.564 & 3.70 & 15.97 & 0.10 & 6.78 & 0.45 & 2.97 & 15.96 \\
3 & KIVI-style & 0.741 & 0.14 & 0.02 & 0.08 & 0.08 & 0.14 & -- & 0.16 \\
3 & AKVQ-style & 0.625 & 2.28 & 11.94 & 0.03 & 4.77 & 0.09 & -- & 11.83 \\
\bottomrule
\end{tabular}
\end{table}

\subsection{The image-sensitivity increment}
Output sensitivity is already a strong selector: at $2$-bit min--max it recovers
$13.34$ points over uniform, versus $1.39$ for random head selection
(Table~\ref{tab:fullkv_selectors}). Adding image sensitivity raises recovery to
$14.37$ points, an additional $1.03$ points averaged over all six tasks.
Image sensitivity alone recovers $2.57$ points.

On POPE with $2$-bit min--max, HeadGuard improves accuracy over output-only head
selection by $1.8$--$2.9$ percentage points for five of the eight models
(Table~\ref{tab:pope2}). The $95\%$ confidence intervals for these five gains
exclude zero.
We compute these intervals using $10{,}000$ paired image-cluster bootstrap
resamples of the full eligible POPE evaluation set.
Each resample compares the two methods' saved predictions on the same questions.
The heads protected by each method are chosen before evaluation and are not
selected again during bootstrapping (Appendix~\ref{app:uncertainty}).
At $3$ bits, hybrid and output-only selection
achieve comparable six-task min--max recovery: $15.96$ and $15.97$ points,
respectively.

\subsection{Keys-only protection retains substantial recovery}
\label{sec:keys}
Protecting only the selected image \emph{keys} in BF16, while quantizing their
values, retains substantial accuracy recovery. It gives
mean $2$-bit gains of $14.56$ points on min--max, $13.95$ on AKVQ-style, and $0.57$
on KIVI-style (Table~\ref{tab:keys_summary}).
Min--max Qwen2.5-VL improves from $0.366$ to $0.751$, while
both Gemma models show little recovery (per-model results in Table~\ref{tab:keys}).

\begin{table}[!htbp]
\centering
\caption{\textbf{K-only results across classification and generation, $2$ bits.}
Uniform has no protection; HG-K is keys-only HeadGuard at $K\approx N_{\rm KV}/8$.
Each metric averages all eight models equally; six-task accuracy also weights
the tasks equally. Entries are scores, not gains, using the metrics in
Section~\ref{sec:protocol}.}
\label{tab:keys_summary}
\setlength{\tabcolsep}{7pt}
\begin{tabular}{@{}l*{6}{c}@{}}
\toprule
 & \multicolumn{2}{c}{Six-task accuracy} & \multicolumn{2}{c}{TextVQA} & \multicolumn{2}{c}{Caption fidelity} \\
\cmidrule(lr){2-3}\cmidrule(lr){4-5}\cmidrule(lr){6-7}
Base & Uniform & HG-K & Uniform & HG-K & Uniform & HG-K \\
\midrule
min--max & 0.434 & 0.580 & 0.145 & 0.357 & 0.167 & 0.313 \\
KIVI-style & 0.688 & 0.693 & 0.554 & 0.556 & 0.471 & 0.493 \\
AKVQ-style & 0.588 & 0.727 & 0.392 & 0.630 & 0.410 & 0.555 \\
\midrule
BF16 reference & \multicolumn{2}{c}{0.745} & \multicolumn{2}{c}{0.680} & \multicolumn{2}{c}{1.000} \\
\bottomrule
\end{tabular}
\end{table}

For $2$-bit min--max, we estimate image-cache storage assuming equal
$128$-dimensional K/V heads and a $1/8$ protected fraction.
Including $32$ bits of quantization metadata per quantized key or value vector,
the estimates are $2.25$, $3.97$, and $3.11$
bits per element for uniform quantization, K+V protection, and K-only protection,
respectively.
For the same selected heads, K-only leaves their image values quantized, halving
the additional storage required for protection and using about $22\%$ less modeled
image-cache storage than K+V.
These estimates describe the storage needed for the image-token KV cache, rather
than measured peak memory usage during end-to-end inference.

At $2$ bits, K-only and K+V protection achieve similar mean accuracies across
the six discriminative tasks and eight models for each base quantizer.

\subsection{Protection can also help generation}
\label{sec:generation}
We also test generated answers on TextVQA and image captions. K+V protection tests
cover all eight models and three base quantizers at $2$ and $3$ bits.
With $2$-bit min--max, K+V HeadGuard raises Qwen2.5-VL's TextVQA score from
$0.015$ to $0.687$. With $3$-bit min--max, Qwen2-VL's score rises from $0.022$
to $0.787$, close to its BF16 score of $0.790$. Even with K+V HeadGuard, both
Gemma models score poorly on TextVQA under $2$-bit min--max quantization.
For captioning with $2$-bit min--max, HeadGuard raises the eight-model mean
ROUGE-L against BF16 output from $0.171$ to $0.312$.

Keys-only protection is tested on all three bases at $2$ bits. For Qwen2.5-VL,
it raises the TextVQA score from $0.009$ to $0.690$ on min--max, but lowers it
from $0.733$ to $0.724$ on KIVI-style. Protection can therefore help generated
answers, but not for every model and base.
Table~\ref{tab:keys_summary} summarizes K-only results across all eight models.
Appendix~\ref{app:generation} reports all results.

\subsection{Ablation: budget, precision, and calibration}
\label{sec:ablation}
We compare head budgets, precision, and calibration on capped task subsets.
For the precision and calibration comparisons, K+V protection is evaluated at
all four head budgets ($1/16$, $1/12$, $1/8$, and $1/4$).
Uniform baselines protect no heads.
At the $1/8$ budget, Appendix~\ref{app:capped_sweep} reports six-task accuracy and
Appendix~\ref{app:capped_generation} reports TextVQA and captioning results.
Appendix~\ref{app:additional_budgets} covers the other three head budgets.

\begin{figure}[t]
\centering
\includegraphics[width=0.88\linewidth]{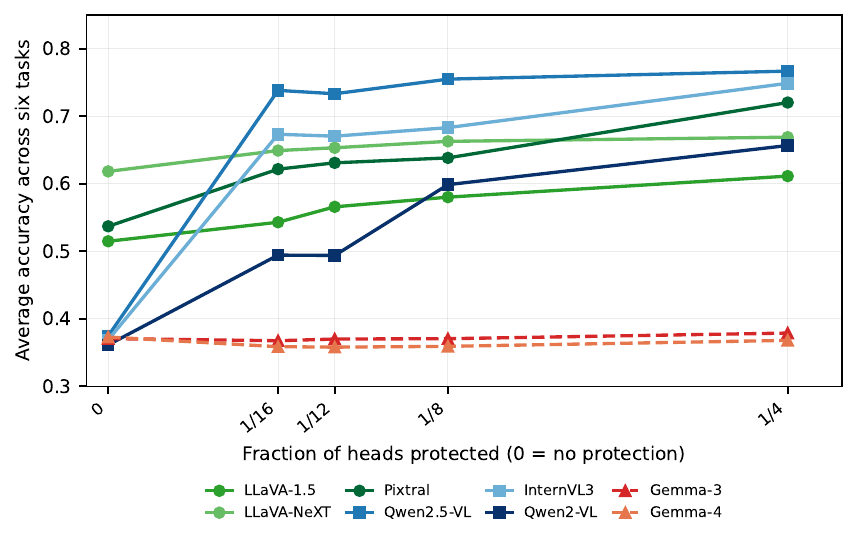}
\caption{\textbf{Accuracy as more heads are protected.}
Selected image keys and values stay in BF16; the rest use $2$-bit min--max.
Each line shows one model's average accuracy across six tasks (POPE, MME,
MMBench, ScienceQA, MMMU, and SEED-Bench), using up to
$1000$ questions per task ($900$ for POPE; $839$ for MMMU).
HeadGuard's ranking is fixed using POPE calibration.
Zero means no protection; fractions are rounded to whole head counts.
The vertical axis shows accuracy, not gains. More protection helps some models,
but both Gemma models remain close to their unprotected scores under this ranking.}
\label{fig:budget}
\end{figure}

Figure~\ref{fig:budget} uses all six tasks of the capped min--max sweep.
Larger protected fractions improve Qwen2-VL and InternVL3. For Gemma,
increasing min--max precision from $2$ to $3$ bits is more effective than
expanding the tested $2$-bit head budget up to $1/4$.

Increasing uniform min--max precision from $2$ to $3$ bits raises Gemma-3's
six-task mean from $0.371$ to $0.656$ and Gemma-4's from $0.373$ to $0.708$.
For Qwen2-VL, uniform $4$-bit min--max still gives $0.366$, versus $0.735$ with
protection at the one-eighth head budget (Table~\ref{tab:sweep_pope}).

At the one-eighth head budget, K+V HeadGuard on min--max raises the
six-task, eight-model mean accuracy over uniform by $14.11$ versus $12.65$
percentage points at $2$ bits and $15.79$ versus $14.39$ at $3$ bits,
for POPE versus MMBench calibration, respectively (Tables~\ref{tab:sweep_pope}
and~\ref{tab:sweep_mmbench}). The image-sensitivity contribution varies
between conditions.

\section{Discussion and Conclusion}
\label{sec:disc}
HeadGuard shows that protecting a small subset of image KV heads can substantially
recover low-bit VLM accuracy without replacing the underlying quantizer. Its
offline sensitivity scores produce a fixed protection mask that is reused across
tasks, retaining all tokens and requiring neither weight updates nor inference-time
gradients. Recovery is particularly strong for Qwen and InternVL on weak bases.
Output sensitivity explains most recovery, while the additional benefit of image
sensitivity depends on the model and calibration.

The benefits extend beyond classification to generated answers and caption
fidelity to BF16 outputs. Keys-only protection retains substantial recovery at a
lower modeled storage cost, offering a more economical protection option. These
findings suggest a practical strategy: choose a strong base quantizer, then use
selective head protection only when its accuracy benefit justifies the added
modeled storage cost. Overall, HeadGuard
demonstrates that a small, targeted high-precision allocation can improve the
accuracy--storage trade-off of low-bit image KV quantization.

\textbf{Scope and limitations.} Results use fixed image-size limits, custom
scoring, and simplified KIVI-style and AKVQ-style bases. We compare POPE and
MMBench calibration, each with $100$ examples, but do not test larger calibration
sets. More calibration examples could further improve head selection and accuracy
recovery, but this possibility remains untested.
Model comparisons do not isolate architectural causes of quantization sensitivity. The main experiments
simulate quantization and estimate storage; end-to-end memory use and speed
with optimized packed caches remain future work.
Because text KV remains in BF16, total-cache compression diminishes as the text
fraction grows; long-form and multi-turn generation are not evaluated.

\section*{Reproducibility Statement}
Full implementation details are provided in the anonymous code supplement,
including experiment-specific runners, evaluation and scoring code, frozen
calibration scores, rounded reference results, aggregation/bootstrap code,
pinned dependencies, and unit tests for the custom TextVQA and caption-fidelity
metrics. Sections~\ref{sec:method} and~\ref{sec:protocol} specify the method,
data splits, scoring, and averaging conventions. Data shuffling uses a fixed seed,
and the released analysis code reproduces the reported aggregates from the
bundled reference results.

Experiments were conducted on NVIDIA H100 and A100 GPUs.
Within each experimental protocol---main K+V evaluation, K-only evaluation,
and the capped budget sweep---the same GPU was used for all models and measurements.
HeadGuard's gains are measured within each evaluation by comparing the same model
and KV-cache quantization method with and without selective head protection.
Comparisons of absolute accuracy across protocols describe trends rather than
quantify HeadGuard's gains.

\section*{Ethics Statement}
The study uses existing public models and benchmarks and collects no new human
data. Model and dataset licenses and privacy limitations still apply. Quantized
VLMs can hallucinate; improved benchmark accuracy does not establish safety in
high-stakes deployment.

\section*{AI Use Statement}
The authors used generative AI assistants (ChatGPT and Claude) for a task with
required disclosure: writing software code that implements methods (scripts for
evaluation, results aggregation, and figure rendering). We additionally used
these tools for recommended-disclosure tasks: editing software code, creating
figures, drafting parts of the manuscript (including descriptions of results),
editing for grammar and readability, and formatting references.
Generative AI was not used to generate synthetic or experimental data, develop
theoretical results or proofs, propose or refine hypotheses, or design the
methodology. All code, numerical results, figures, scientific claims, conclusions,
and their interpretation were reviewed, validated, and finalized by the authors,
who take full responsibility for the content of this article.

\label{main:end}

\clearpage
\appendix
\section*{Appendix}
\label{app:start}
Detailed results follow the protocols in Section~\ref{sec:protocol}.

\section{Key-and-Value Protection at Two and Three Bits}
\label{app:fullkv}
Tables~\ref{tab:fullkv_tasks2} and~\ref{tab:fullkv_tasks3} give per-task min--max
results. U denotes uniform quantization; HG denotes HeadGuard.

\begin{table}[!htbp]
\centering
\caption{\textbf{K+V protection, min--max at $2$ bits}, all six tasks. Each pair is uniform/HG;
$K\approx N_{\rm KV}/8$. Counts are $8892/2366/4321/2009/839/14225$ in column order.}
\label{tab:fullkv_tasks2}
\setlength{\tabcolsep}{2.4pt}
\begin{tabular}{@{}l*{12}{c}@{}}
\toprule
 & \multicolumn{2}{c}{POPE} & \multicolumn{2}{c}{MME} & \multicolumn{2}{c}{MMBench} & \multicolumn{2}{c}{ScienceQA} & \multicolumn{2}{c}{MMMU} & \multicolumn{2}{c}{SEED} \\
Model & U & HG & U & HG & U & HG & U & HG & U & HG & U & HG \\
\midrule
LLaVA-1.5 & 0.737 & 0.832 & 0.581 & 0.672 & 0.513 & 0.594 & 0.509 & 0.562 & 0.324 & 0.313 & 0.412 & 0.503 \\
LLaVA-NeXT & 0.832 & 0.857 & 0.745 & 0.750 & 0.664 & 0.711 & 0.585 & 0.634 & 0.331 & 0.336 & 0.568 & 0.653 \\
Qwen2.5-VL & 0.496 & 0.866 & 0.500 & 0.821 & 0.266 & 0.831 & 0.361 & 0.787 & 0.278 & 0.449 & 0.260 & 0.727 \\
Qwen2-VL & 0.510 & 0.676 & 0.504 & 0.598 & 0.271 & 0.713 & 0.349 & 0.678 & 0.284 & 0.347 & 0.261 & 0.575 \\
Gemma-3 & 0.517 & 0.509 & 0.533 & 0.514 & 0.273 & 0.268 & 0.365 & 0.359 & 0.280 & 0.286 & 0.263 & 0.266 \\
Gemma-4 & 0.500 & 0.510 & 0.493 & 0.486 & 0.270 & 0.272 & 0.367 & 0.386 & 0.288 & 0.271 & 0.251 & 0.255 \\
Pixtral & 0.664 & 0.810 & 0.592 & 0.706 & 0.653 & 0.728 & 0.632 & 0.729 & 0.312 & 0.360 & 0.376 & 0.516 \\
InternVL3 & 0.498 & 0.856 & 0.506 & 0.712 & 0.277 & 0.746 & 0.347 & 0.769 & 0.292 & 0.402 & 0.253 & 0.640 \\
\bottomrule
\end{tabular}
\end{table}

\begin{table}[!htbp]
\centering
\caption{\textbf{K+V protection, min--max at $3$ bits}, same complete six-task block and
column conventions as Table~\ref{tab:fullkv_tasks2}.}
\label{tab:fullkv_tasks3}
\setlength{\tabcolsep}{2.4pt}
\begin{tabular}{@{}l*{12}{c}@{}}
\toprule
 & \multicolumn{2}{c}{POPE} & \multicolumn{2}{c}{MME} & \multicolumn{2}{c}{MMBench} & \multicolumn{2}{c}{ScienceQA} & \multicolumn{2}{c}{MMMU} & \multicolumn{2}{c}{SEED} \\
Model & U & HG & U & HG & U & HG & U & HG & U & HG & U & HG \\
\midrule
LLaVA-1.5 & 0.840 & 0.844 & 0.782 & 0.782 & 0.679 & 0.680 & 0.611 & 0.615 & 0.356 & 0.355 & 0.579 & 0.577 \\
LLaVA-NeXT & 0.866 & 0.860 & 0.770 & 0.754 & 0.710 & 0.713 & 0.655 & 0.658 & 0.356 & 0.361 & 0.646 & 0.650 \\
Qwen2.5-VL & 0.511 & 0.881 & 0.501 & 0.855 & 0.268 & 0.856 & 0.347 & 0.836 & 0.281 & 0.492 & 0.255 & 0.763 \\
Qwen2-VL & 0.495 & 0.882 & 0.502 & 0.806 & 0.267 & 0.785 & 0.350 & 0.772 & 0.277 & 0.414 & 0.262 & 0.714 \\
Gemma-3 & 0.794 & 0.818 & 0.814 & 0.843 & 0.672 & 0.734 & 0.696 & 0.734 & 0.418 & 0.445 & 0.509 & 0.583 \\
Gemma-4 & 0.794 & 0.815 & 0.730 & 0.740 & 0.793 & 0.807 & 0.763 & 0.778 & 0.458 & 0.428 & 0.691 & 0.705 \\
Pixtral & 0.859 & 0.860 & 0.829 & 0.829 & 0.829 & 0.834 & 0.872 & 0.874 & 0.477 & 0.478 & 0.720 & 0.723 \\
InternVL3 & 0.527 & 0.896 & 0.513 & 0.838 & 0.275 & 0.862 & 0.354 & 0.942 & 0.273 & 0.534 & 0.262 & 0.742 \\
\bottomrule
\end{tabular}
\end{table}

\section{Keys-only Discriminative Evaluation}
Table~\ref{tab:keys} gives the per-model means for keys-only protection on the
full eligible discriminative splits, separately from K+V evaluation.

\begin{table}[!htbp]
\centering
\caption{\textbf{K-only evaluation at $2$ bits, all six discriminative tasks.}
$K\approx N_{\rm KV}/8$.
Each pair gives accuracy without protection (Uniform) and with keys-only
HeadGuard (HG-K), not gains.}
\label{tab:keys}
\setlength{\tabcolsep}{5pt}
\begin{tabular}{@{}lccccccc@{}}
\toprule
 & \multicolumn{2}{c}{min--max} & \multicolumn{2}{c}{AKVQ-style} & \multicolumn{2}{c}{KIVI-style} & \\
\cmidrule(lr){2-3}\cmidrule(lr){4-5}\cmidrule(lr){6-7}
Model & Uniform & HG-K & Uniform & HG-K & Uniform & HG-K & BF16 \\
\midrule
LLaVA-1.5 & 0.508 & 0.573 & 0.633 & 0.640 & 0.639 & 0.639 & 0.643 \\
LLaVA-NeXT & 0.615 & 0.660 & 0.661 & 0.664 & 0.655 & 0.663 & 0.664 \\
Qwen2.5-VL & 0.366 & 0.751 & 0.395 & 0.774 & 0.777 & 0.782 & 0.788 \\
Qwen2-VL & 0.361 & 0.605 & 0.364 & 0.707 & 0.717 & 0.720 & 0.732 \\
Gemma-3 & 0.364 & 0.365 & 0.695 & 0.725 & 0.732 & 0.741 & 0.755 \\
Gemma-4 & 0.364 & 0.368 & 0.783 & 0.784 & 0.433 & 0.435 & 0.786 \\
Pixtral & 0.536 & 0.634 & 0.758 & 0.763 & 0.744 & 0.755 & 0.770 \\
InternVL3 & 0.359 & 0.684 & 0.414 & 0.761 & 0.804 & 0.813 & 0.825 \\
\bottomrule
\end{tabular}
\end{table}

\clearpage
\section{POPE Uncertainty in Full Evaluation}
\label{app:uncertainty}
POPE is used here as an evaluation benchmark, not only for calibration.
The bootstrap uses saved predictions on the full eligible POPE evaluation set,
excluding calibration and diagnostic questions.

Tables~\ref{tab:pope2}--\ref{tab:keys_ci} use the same paired image-cluster
percentile bootstrap for $95\%$ intervals: $10{,}000$ draws with seed $0$,
keeping each image's questions together.
Entries are HeadGuard's accuracy gains over the named baseline, in percentage
points, with pointwise $95\%$ confidence intervals in brackets.
Positive values favor HeadGuard; negative values favor the baseline.
An interval spanning zero does not establish a clear difference.

\begin{table}[!htbp]
\centering
\caption{\textbf{K+V protection, $2$-bit min--max POPE}: paired image-cluster
$95\%$ intervals, $K\approx N_{\rm KV}/8$.
Gains and intervals are in percentage points, not absolute accuracies.}
\label{tab:pope2}
\begin{tabular}{@{}lcc@{}}
\toprule
Model & \shortstack{HeadGuard gain\\over Uniform [CI]} & \shortstack{HeadGuard gain\\over Output-only [CI]} \\
\midrule
LLaVA-1.5 & $+9.5\,[+8.1,+10.8]$ & $+1.8\,[+0.9,+2.7]$ \\
LLaVA-NeXT & $+2.5\,[+1.4,+3.6]$ & $+2.6\,[+1.7,+3.4]$ \\
Qwen2.5-VL & $+36.9\,[+35.5,+38.3]$ & $+2.9\,[+2.0,+3.9]$ \\
Qwen2-VL & $+16.5\,[+14.7,+18.4]$ & $+0.3\,[-1.1,+1.7]$ \\
Gemma-3 & $-0.8\,[-2.6,+1.0]$ & $-0.8\,[-2.4,+0.8]$ \\
Gemma-4 & $+1.0\,[-0.7,+2.7]$ & $+0.7\,[-0.9,+2.2]$ \\
Pixtral & $+14.6\,[+13.1,+16.1]$ & $+2.3\,[+1.0,+3.5]$ \\
InternVL3 & $+35.8\,[+34.6,+37.1]$ & $+1.9\,[+0.8,+2.9]$ \\
\bottomrule
\end{tabular}
\end{table}

\begin{table}[!htbp]
\centering
\caption{\textbf{K+V protection, $3$-bit min--max POPE}: same paired
image-cluster bootstrap and head budget as Table~\ref{tab:pope2}.
Gains and intervals are in percentage points, not absolute accuracies.}
\label{tab:pope3}
\begin{tabular}{@{}lcc@{}}
\toprule
Model & \shortstack{HeadGuard gain\\over Uniform [CI]} & \shortstack{HeadGuard gain\\over Output-only [CI]} \\
\midrule
LLaVA-1.5 & $+0.4\,[-0.1,+0.9]$ & $0.0\,[-0.4,+0.5]$ \\
LLaVA-NeXT & $-0.6\,[-1.0,-0.2]$ & $-0.4\,[-0.7,-0.1]$ \\
Qwen2.5-VL & $+37.0\,[+35.8,+38.2]$ & $-0.3\,[-0.8,+0.2]$ \\
Qwen2-VL & $+38.7\,[+37.3,+40.0]$ & $+0.1\,[-0.3,+0.5]$ \\
Gemma-3 & $+2.4\,[+1.4,+3.4]$ & $+0.4\,[-0.5,+1.3]$ \\
Gemma-4 & $+2.1\,[+0.9,+3.4]$ & $-0.9\,[-2.0,+0.2]$ \\
Pixtral & $+0.1\,[-0.4,+0.6]$ & $+0.1\,[-0.3,+0.6]$ \\
InternVL3 & $+36.9\,[+35.5,+38.3]$ & $-0.4\,[-1.0,+0.2]$ \\
\bottomrule
\end{tabular}
\end{table}

\begin{table}[!htbp]
\centering
\caption{\textbf{K-only evaluation, $2$-bit min--max POPE}: same paired
image-cluster bootstrap and head budget as Table~\ref{tab:pope2}.
Gains and intervals are in percentage points, not absolute accuracies.
Contrasts compare selectors within K-only evaluation, not K-only against K+V.}
\label{tab:keys_ci}
\begin{tabular}{@{}lcc@{}}
\toprule
Model & \shortstack{HeadGuard gain\\over Uniform [CI]} & \shortstack{HeadGuard gain\\over Output-only [CI]} \\
\midrule
LLaVA-1.5 & $+8.9\,[+7.5,+10.3]$ & $+0.7\,[-0.2,+1.7]$ \\
LLaVA-NeXT & $+3.0\,[+1.9,+4.1]$ & $+2.7\,[+1.8,+3.4]$ \\
Qwen2.5-VL & $+36.5\,[+35.1,+38.0]$ & $+3.1\,[+2.1,+4.0]$ \\
Qwen2-VL & $+17.8\,[+16.0,+19.6]$ & $-1.1\,[-2.6,+0.4]$ \\
Gemma-3 & $-0.7\,[-2.4,+1.0]$ & $+0.6\,[-1.0,+2.2]$ \\
Gemma-4 & $0.0\,[-1.5,+1.6]$ & $-0.6\,[-2.0,+0.9]$ \\
Pixtral & $+13.6\,[+12.1,+15.0]$ & $+2.2\,[+0.9,+3.6]$ \\
InternVL3 & $+34.9\,[+33.6,+36.1]$ & $+1.6\,[+0.5,+2.8]$ \\
\bottomrule
\end{tabular}
\end{table}

\section{Generative Results}
\label{app:generation}
K+V evaluation reports all model--base pairs for each generative task
at both precisions. K-only evaluation reports TextVQA and caption
pairs. TextVQA uses $1000$ rows and captioning $200$; scores are descriptive
and use the custom metrics defined above.
U denotes no head protection, HG denotes HeadGuard, and HG-K denotes keys-only
HeadGuard. Entries are task scores, not gains: higher is better for TextVQA;
higher caption fidelity means closer agreement with BF16 output, not necessarily
better captions.

\begin{table}[!htbp]
\centering
\caption{\textbf{K+V protection on TextVQA, $2$ bits.} Uniform/HG pairs;
all eight models are retained.}
\label{tab:fullkv_textvqa2}
\begin{tabular}{@{}lccccccc@{}}
\toprule
 & \multicolumn{2}{c}{min--max} & \multicolumn{2}{c}{AKVQ-style} & \multicolumn{2}{c}{KIVI-style} & \\
Model & U & HG & U & HG & U & HG & BF16 \\
\midrule
LLaVA-1.5 & 0.313 & 0.382 & 0.460 & 0.470 & 0.459 & 0.470 & 0.476 \\
LLaVA-NeXT & 0.497 & 0.552 & 0.554 & 0.560 & 0.558 & 0.547 & 0.581 \\
Qwen2.5-VL & 0.015 & 0.687 & 0.039 & 0.744 & 0.716 & 0.731 & 0.775 \\
Qwen2-VL & 0.064 & 0.412 & 0.056 & 0.718 & 0.717 & 0.730 & 0.790 \\
Gemma-3 & 0.008 & 0.012 & 0.579 & 0.618 & 0.662 & 0.689 & 0.706 \\
Gemma-4 & 0.002 & 0.004 & 0.671 & 0.678 & 0.007 & 0.021 & 0.685 \\
Pixtral & 0.229 & 0.405 & 0.714 & 0.719 & 0.669 & 0.688 & 0.729 \\
InternVL3 & 0.022 & 0.443 & 0.193 & 0.597 & 0.638 & 0.653 & 0.706 \\
\bottomrule
\end{tabular}
\end{table}

\begin{table}[!htbp]
\centering
\caption{\textbf{K+V protection on TextVQA, $3$ bits.} Same conventions as
Table~\ref{tab:fullkv_textvqa2}.}
\label{tab:fullkv_textvqa3}
\begin{tabular}{@{}lccccccc@{}}
\toprule
 & \multicolumn{2}{c}{min--max} & \multicolumn{2}{c}{AKVQ-style} & \multicolumn{2}{c}{KIVI-style} & \\
Model & U & HG & U & HG & U & HG & BF16 \\
\midrule
LLaVA-1.5 & 0.480 & 0.479 & 0.475 & 0.475 & 0.474 & 0.480 & 0.476 \\
LLaVA-NeXT & 0.583 & 0.575 & 0.580 & 0.571 & 0.576 & 0.583 & 0.581 \\
Qwen2.5-VL & 0.000 & 0.756 & 0.003 & 0.760 & 0.761 & 0.771 & 0.775 \\
Qwen2-VL & 0.022 & 0.787 & 0.253 & 0.785 & 0.775 & 0.783 & 0.790 \\
Gemma-3 & 0.556 & 0.615 & 0.702 & 0.694 & 0.707 & 0.701 & 0.706 \\
Gemma-4 & 0.308 & 0.344 & 0.679 & 0.683 & 0.510 & 0.541 & 0.685 \\
Pixtral & 0.728 & 0.725 & 0.722 & 0.728 & 0.725 & 0.722 & 0.729 \\
InternVL3 & 0.031 & 0.672 & 0.301 & 0.705 & 0.704 & 0.698 & 0.706 \\
\bottomrule
\end{tabular}
\end{table}

\begin{table}[!htbp]
\centering
\caption{\textbf{K+V protection for captioning, $2$ bits.} ROUGE-L overlap with BF16
output, not human-caption quality. Each pair is uniform/HG on $200$ examples.}
\label{tab:fullkv_caption2}
\begin{tabular}{@{}lccccccc@{}}
\toprule
 & \multicolumn{2}{c}{min--max} & \multicolumn{2}{c}{AKVQ-style} & \multicolumn{2}{c}{KIVI-style} & \\
Model & U & HG & U & HG & U & HG & BF16 \\
\midrule
LLaVA-1.5 & 0.370 & 0.477 & 0.631 & 0.661 & 0.627 & 0.664 & 1.000 \\
LLaVA-NeXT & 0.492 & 0.514 & 0.663 & 0.681 & 0.557 & 0.606 & 1.000 \\
Qwen2.5-VL & 0.004 & 0.375 & 0.118 & 0.454 & 0.452 & 0.476 & 1.000 \\
Qwen2-VL & 0.058 & 0.280 & 0.014 & 0.466 & 0.374 & 0.392 & 1.000 \\
Gemma-3 & 0.077 & 0.136 & 0.523 & 0.548 & 0.570 & 0.613 & 1.000 \\
Gemma-4 & 0.037 & 0.049 & 0.749 & 0.738 & 0.113 & 0.125 & 1.000 \\
Pixtral & 0.264 & 0.322 & 0.616 & 0.691 & 0.582 & 0.639 & 1.000 \\
InternVL3 & 0.065 & 0.344 & 0.159 & 0.393 & 0.489 & 0.504 & 1.000 \\
\bottomrule
\end{tabular}
\end{table}

\begin{table}[!htbp]
\centering
\caption{\textbf{K+V protection for captioning, $3$ bits.} Same scope and conventions
as Table~\ref{tab:fullkv_caption2}.}
\label{tab:fullkv_caption3}
\begin{tabular}{@{}lccccccc@{}}
\toprule
 & \multicolumn{2}{c}{min--max} & \multicolumn{2}{c}{AKVQ-style} & \multicolumn{2}{c}{KIVI-style} & \\
Model & U & HG & U & HG & U & HG & BF16 \\
\midrule
LLaVA-1.5 & 0.717 & 0.748 & 0.804 & 0.828 & 0.783 & 0.806 & 1.000 \\
LLaVA-NeXT & 0.722 & 0.796 & 0.815 & 0.846 & 0.806 & 0.840 & 1.000 \\
Qwen2.5-VL & 0.009 & 0.427 & 0.025 & 0.468 & 0.625 & 0.611 & 1.000 \\
Qwen2-VL & 0.025 & 0.581 & 0.022 & 0.651 & 0.617 & 0.642 & 1.000 \\
Gemma-3 & 0.536 & 0.556 & 0.692 & 0.722 & 0.706 & 0.755 & 1.000 \\
Gemma-4 & 0.383 & 0.402 & 0.817 & 0.814 & 0.473 & 0.488 & 1.000 \\
Pixtral & 0.773 & 0.790 & 0.829 & 0.864 & 0.812 & 0.818 & 1.000 \\
InternVL3 & 0.054 & 0.457 & 0.140 & 0.518 & 0.668 & 0.666 & 1.000 \\
\bottomrule
\end{tabular}
\end{table}

\begin{table}[!htbp]
\centering
\caption{\textbf{K-only TextVQA evaluation, $2$ bits.}
Each pair is uniform/HG-K on $1000$ questions.}
\label{tab:keys_textvqa}
\begin{tabular}{@{}lccccccc@{}}
\toprule
 & \multicolumn{2}{c}{min--max} & \multicolumn{2}{c}{AKVQ-style} & \multicolumn{2}{c}{KIVI-style} & \\
Model & U & HG-K & U & HG-K & U & HG-K & BF16 \\
\midrule
LLaVA-1.5 & 0.312 & 0.379 & 0.471 & 0.468 & 0.460 & 0.452 & 0.475 \\
LLaVA-NeXT & 0.518 & 0.531 & 0.562 & 0.558 & 0.557 & 0.553 & 0.578 \\
Qwen2.5-VL & 0.009 & 0.690 & 0.025 & 0.750 & 0.733 & 0.724 & 0.771 \\
Qwen2-VL & 0.050 & 0.425 & 0.060 & 0.707 & 0.705 & 0.718 & 0.791 \\
Gemma-3 & 0.009 & 0.018 & 0.575 & 0.603 & 0.665 & 0.683 & 0.705 \\
Gemma-4 & 0.002 & 0.003 & 0.655 & 0.658 & 0.010 & 0.008 & 0.685 \\
Pixtral & 0.230 & 0.355 & 0.708 & 0.712 & 0.673 & 0.677 & 0.727 \\
InternVL3 & 0.028 & 0.454 & 0.078 & 0.584 & 0.626 & 0.634 & 0.710 \\
\bottomrule
\end{tabular}
\end{table}

\begin{table}[!htbp]
\centering
\caption{\textbf{K-only caption evaluation, $2$ bits.}
$200$ examples per cell; each pair is uniform/HG-K.}
\label{tab:keys_caption}
\begin{tabular}{@{}lccccccc@{}}
\toprule
 & \multicolumn{2}{c}{min--max} & \multicolumn{2}{c}{AKVQ-style} & \multicolumn{2}{c}{KIVI-style} & \\
Model & U & HG-K & U & HG-K & U & HG-K & BF16 \\
\midrule
LLaVA-1.5 & 0.373 & 0.472 & 0.646 & 0.647 & 0.638 & 0.645 & 1.000 \\
LLaVA-NeXT & 0.460 & 0.533 & 0.626 & 0.664 & 0.551 & 0.603 & 1.000 \\
Qwen2.5-VL & 0.005 & 0.368 & 0.095 & 0.426 & 0.451 & 0.467 & 1.000 \\
Qwen2-VL & 0.053 & 0.300 & 0.015 & 0.433 & 0.380 & 0.395 & 1.000 \\
Gemma-3 & 0.073 & 0.132 & 0.518 & 0.534 & 0.580 & 0.584 & 1.000 \\
Gemma-4 & 0.037 & 0.042 & 0.667 & 0.662 & 0.096 & 0.106 & 1.000 \\
Pixtral & 0.262 & 0.319 & 0.616 & 0.673 & 0.578 & 0.616 & 1.000 \\
InternVL3 & 0.071 & 0.339 & 0.099 & 0.397 & 0.491 & 0.529 & 1.000 \\
\bottomrule
\end{tabular}
\end{table}

\clearpage
\section{Capped Budget Sweep}
\label{app:capped_sweep}

At $2$-bit min--max with the one-eighth head budget, HeadGuard improves mean accuracy over uniform by
$14.11$ percentage points under POPE calibration and $12.65$ under MMBench calibration
(Tables~\ref{tab:sweep_pope} and~\ref{tab:sweep_mmbench}).

In this section and Appendix~\ref{app:capped_generation}, U = uniform;
Out = output-only selection; HG = hybrid HeadGuard.
Out uses only output sensitivity, whereas HG combines image-sensitivity and
output-sensitivity scores. Entries are accuracies or task scores, not gains.
Mean weights all eight models equally within each metric, base, and precision.

\begin{table}[!htbp]
\centering
\caption{\textbf{Capped sweep with POPE calibration.} Each score averages
POPE, MME, MMBench, ScienceQA, MMMU, and SEED, using up to $1000$ questions per task
($900$ for POPE; $839$ for MMMU). K+V protection uses $K=\operatorname{round}(N_{\rm KV}/8)$.
U = uniform; Out = output-only selection; HG = hybrid HeadGuard. Mean averages
all eight models. Values are accuracies, not
gains; gains in the main text are calculated before final rounding.}
\label{tab:sweep_pope}
\setlength{\tabcolsep}{4.25pt}
\renewcommand{\arraystretch}{1.12}
\begin{tabular}{@{}lc*{10}{c}@{}}
\toprule
 & & \multicolumn{3}{c}{min--max} & \multicolumn{3}{c}{AKVQ-style} & \multicolumn{3}{c}{KIVI-style} & \\
\cmidrule(lr){3-5}\cmidrule(lr){6-8}\cmidrule(lr){9-11}
Model & Bits & U & Out & HG & U & Out & HG & U & Out & HG & BF16 \\
\midrule
LLaVA-1.5 & 2 & 0.515 & 0.560 & 0.580 & 0.637 & 0.633 & 0.641 & 0.638 & 0.636 & 0.636 & 0.643 \\
LLaVA-1.5 & 3 & 0.640 & 0.642 & 0.641 & 0.642 & 0.642 & 0.645 & 0.644 & 0.644 & 0.645 & 0.643 \\
LLaVA-1.5 & 4 & 0.643 & 0.644 & 0.642 & 0.644 & 0.644 & 0.642 & 0.642 & 0.644 & 0.643 & 0.643 \\
\addlinespace[2pt]
LLaVA-NeXT & 2 & 0.618 & 0.648 & 0.663 & 0.663 & 0.667 & 0.668 & 0.656 & 0.665 & 0.665 & 0.666 \\
LLaVA-NeXT & 3 & 0.667 & 0.669 & 0.667 & 0.668 & 0.666 & 0.665 & 0.666 & 0.665 & 0.665 & 0.666 \\
LLaVA-NeXT & 4 & 0.667 & 0.668 & 0.666 & 0.665 & 0.668 & 0.665 & 0.667 & 0.665 & 0.665 & 0.666 \\
\addlinespace[2pt]
Qwen2.5-VL & 2 & 0.375 & 0.722 & 0.755 & 0.400 & 0.779 & 0.781 & 0.782 & 0.784 & 0.785 & 0.792 \\
Qwen2.5-VL & 3 & 0.369 & 0.788 & 0.782 & 0.376 & 0.792 & 0.793 & 0.793 & 0.794 & 0.794 & 0.792 \\
Qwen2.5-VL & 4 & 0.417 & 0.795 & 0.791 & 0.569 & 0.792 & 0.791 & 0.793 & 0.793 & 0.794 & 0.792 \\
\addlinespace[2pt]
Qwen2-VL & 2 & 0.361 & 0.584 & 0.599 & 0.365 & 0.702 & 0.703 & 0.719 & 0.715 & 0.721 & 0.737 \\
Qwen2-VL & 3 & 0.361 & 0.729 & 0.730 & 0.497 & 0.731 & 0.731 & 0.734 & 0.733 & 0.734 & 0.737 \\
Qwen2-VL & 4 & 0.366 & 0.733 & 0.735 & 0.528 & 0.735 & 0.735 & 0.734 & 0.737 & 0.737 & 0.737 \\
\addlinespace[2pt]
Gemma-3 & 2 & 0.371 & 0.369 & 0.371 & 0.701 & 0.724 & 0.733 & 0.738 & 0.748 & 0.747 & 0.758 \\
Gemma-3 & 3 & 0.656 & 0.679 & 0.695 & 0.755 & 0.757 & 0.761 & 0.757 & 0.759 & 0.759 & 0.758 \\
Gemma-3 & 4 & 0.754 & 0.756 & 0.760 & 0.757 & 0.758 & 0.760 & 0.758 & 0.757 & 0.759 & 0.758 \\
\addlinespace[2pt]
Gemma-4 & 2 & 0.373 & 0.363 & 0.359 & 0.784 & 0.787 & 0.787 & 0.436 & 0.459 & 0.455 & 0.787 \\
Gemma-4 & 3 & 0.708 & 0.722 & 0.711 & 0.785 & 0.784 & 0.785 & 0.750 & 0.756 & 0.759 & 0.787 \\
Gemma-4 & 4 & 0.771 & 0.777 & 0.775 & 0.787 & 0.788 & 0.787 & 0.779 & 0.779 & 0.779 & 0.787 \\
\addlinespace[2pt]
Pixtral & 2 & 0.537 & 0.647 & 0.638 & 0.762 & 0.768 & 0.766 & 0.747 & 0.755 & 0.762 & 0.773 \\
Pixtral & 3 & 0.765 & 0.771 & 0.772 & 0.770 & 0.772 & 0.771 & 0.771 & 0.773 & 0.775 & 0.773 \\
Pixtral & 4 & 0.772 & 0.771 & 0.773 & 0.773 & 0.774 & 0.773 & 0.772 & 0.773 & 0.773 & 0.773 \\
\addlinespace[2pt]
InternVL3 & 2 & 0.369 & 0.688 & 0.683 & 0.415 & 0.783 & 0.760 & 0.803 & 0.808 & 0.813 & 0.823 \\
InternVL3 & 3 & 0.373 & 0.807 & 0.802 & 0.496 & 0.821 & 0.818 & 0.821 & 0.824 & 0.823 & 0.823 \\
InternVL3 & 4 & 0.410 & 0.824 & 0.820 & 0.682 & 0.827 & 0.818 & 0.823 & 0.826 & 0.825 & 0.823 \\
\midrule
Mean & 2 & 0.440 & 0.572 & 0.581 & 0.591 & 0.730 & 0.730 & 0.690 & 0.696 & 0.698 & 0.747 \\
Mean & 3 & 0.567 & 0.726 & 0.725 & 0.623 & 0.746 & 0.746 & 0.742 & 0.743 & 0.744 & 0.747 \\
Mean & 4 & 0.600 & 0.746 & 0.745 & 0.676 & 0.748 & 0.746 & 0.746 & 0.747 & 0.747 & 0.747 \\
\bottomrule
\end{tabular}
\end{table}

\clearpage
\begin{table}[!htbp]
\centering
\caption{\textbf{Capped sweep with MMBench calibration.} Same six evaluation
tasks, one-eighth head budget, and column definitions as Table~\ref{tab:sweep_pope}.
Here MMBench uses $900$ held-out questions, POPE $1000$, and MMMU $839$.
Mean averages all eight models.}
\label{tab:sweep_mmbench}
\setlength{\tabcolsep}{4.25pt}
\renewcommand{\arraystretch}{1.12}
\begin{tabular}{@{}lc*{10}{c}@{}}
\toprule
 & & \multicolumn{3}{c}{min--max} & \multicolumn{3}{c}{AKVQ-style} & \multicolumn{3}{c}{KIVI-style} & \\
\cmidrule(lr){3-5}\cmidrule(lr){6-8}\cmidrule(lr){9-11}
Model & Bits & U & Out & HG & U & Out & HG & U & Out & HG & BF16 \\
\midrule
LLaVA-1.5 & 2 & 0.514 & 0.578 & 0.590 & 0.639 & 0.635 & 0.640 & 0.639 & 0.631 & 0.640 & 0.644 \\
LLaVA-1.5 & 3 & 0.641 & 0.641 & 0.644 & 0.643 & 0.644 & 0.645 & 0.645 & 0.645 & 0.645 & 0.644 \\
LLaVA-1.5 & 4 & 0.644 & 0.644 & 0.646 & 0.646 & 0.647 & 0.644 & 0.643 & 0.646 & 0.645 & 0.644 \\
\addlinespace[2pt]
LLaVA-NeXT & 2 & 0.620 & 0.643 & 0.654 & 0.663 & 0.665 & 0.666 & 0.656 & 0.662 & 0.666 & 0.666 \\
LLaVA-NeXT & 3 & 0.666 & 0.667 & 0.668 & 0.668 & 0.668 & 0.666 & 0.667 & 0.666 & 0.664 & 0.666 \\
LLaVA-NeXT & 4 & 0.668 & 0.666 & 0.667 & 0.666 & 0.666 & 0.668 & 0.668 & 0.667 & 0.666 & 0.666 \\
\addlinespace[2pt]
Qwen2.5-VL & 2 & 0.373 & 0.718 & 0.673 & 0.400 & 0.774 & 0.627 & 0.781 & 0.783 & 0.785 & 0.793 \\
Qwen2.5-VL & 3 & 0.369 & 0.781 & 0.738 & 0.375 & 0.793 & 0.756 & 0.793 & 0.794 & 0.795 & 0.793 \\
Qwen2.5-VL & 4 & 0.415 & 0.792 & 0.765 & 0.569 & 0.792 & 0.765 & 0.793 & 0.794 & 0.792 & 0.793 \\
\addlinespace[2pt]
Qwen2-VL & 2 & 0.361 & 0.594 & 0.513 & 0.364 & 0.700 & 0.610 & 0.718 & 0.721 & 0.720 & 0.737 \\
Qwen2-VL & 3 & 0.361 & 0.730 & 0.660 & 0.498 & 0.734 & 0.710 & 0.734 & 0.732 & 0.734 & 0.737 \\
Qwen2-VL & 4 & 0.366 & 0.736 & 0.629 & 0.528 & 0.736 & 0.723 & 0.734 & 0.736 & 0.737 & 0.737 \\
\addlinespace[2pt]
Gemma-3 & 2 & 0.370 & 0.365 & 0.361 & 0.700 & 0.720 & 0.719 & 0.737 & 0.745 & 0.744 & 0.757 \\
Gemma-3 & 3 & 0.656 & 0.685 & 0.685 & 0.754 & 0.754 & 0.757 & 0.757 & 0.759 & 0.760 & 0.757 \\
Gemma-3 & 4 & 0.754 & 0.754 & 0.758 & 0.757 & 0.759 & 0.760 & 0.758 & 0.759 & 0.759 & 0.757 \\
\addlinespace[2pt]
Gemma-4 & 2 & 0.373 & 0.374 & 0.372 & 0.783 & 0.787 & 0.783 & 0.439 & 0.444 & 0.458 & 0.786 \\
Gemma-4 & 3 & 0.707 & 0.717 & 0.722 & 0.784 & 0.786 & 0.786 & 0.750 & 0.756 & 0.762 & 0.786 \\
Gemma-4 & 4 & 0.770 & 0.775 & 0.772 & 0.786 & 0.787 & 0.787 & 0.777 & 0.779 & 0.777 & 0.786 \\
\addlinespace[2pt]
Pixtral & 2 & 0.537 & 0.639 & 0.661 & 0.761 & 0.762 & 0.768 & 0.746 & 0.756 & 0.760 & 0.773 \\
Pixtral & 3 & 0.765 & 0.767 & 0.772 & 0.769 & 0.770 & 0.773 & 0.770 & 0.769 & 0.773 & 0.773 \\
Pixtral & 4 & 0.772 & 0.772 & 0.772 & 0.773 & 0.772 & 0.772 & 0.772 & 0.772 & 0.773 & 0.773 \\
\addlinespace[2pt]
InternVL3 & 2 & 0.369 & 0.700 & 0.703 & 0.415 & 0.788 & 0.768 & 0.803 & 0.807 & 0.813 & 0.823 \\
InternVL3 & 3 & 0.374 & 0.812 & 0.803 & 0.498 & 0.825 & 0.816 & 0.820 & 0.822 & 0.821 & 0.823 \\
InternVL3 & 4 & 0.413 & 0.825 & 0.816 & 0.682 & 0.825 & 0.818 & 0.822 & 0.823 & 0.824 & 0.823 \\
\midrule
Mean & 2 & 0.439 & 0.576 & 0.566 & 0.591 & 0.729 & 0.698 & 0.690 & 0.693 & 0.698 & 0.747 \\
Mean & 3 & 0.567 & 0.725 & 0.711 & 0.624 & 0.747 & 0.739 & 0.742 & 0.743 & 0.744 & 0.747 \\
Mean & 4 & 0.600 & 0.745 & 0.728 & 0.676 & 0.748 & 0.742 & 0.746 & 0.747 & 0.747 & 0.747 \\
\bottomrule
\end{tabular}
\end{table}

\clearpage
\section{Capped Generative Evaluation}
\label{app:capped_generation}
Tables~\ref{tab:sweep_pope_textvqa}--\ref{tab:sweep_mmbench_caption} report
TextVQA and captioning separately under both calibrations, using $200$ examples
per task and K+V protection at $K=\operatorname{round}(N_{\rm KV}/8)$.
Scores use the custom metrics in Section~\ref{sec:protocol} and are descriptive;
neither generative metric is pooled with the six-task accuracy mean.

\begin{table}[!htbp]
\centering
\caption{\textbf{Capped TextVQA with POPE calibration.} Custom VQA-style scores
on $200$ questions, K+V protection at the one-eighth head budget.
Mean averages all eight models.}
\label{tab:sweep_pope_textvqa}
\setlength{\tabcolsep}{4.25pt}
\renewcommand{\arraystretch}{1.12}
\begin{tabular}{@{}lc*{10}{c}@{}}
\toprule
 & & \multicolumn{3}{c}{min--max} & \multicolumn{3}{c}{AKVQ-style} & \multicolumn{3}{c}{KIVI-style} & \\
\cmidrule(lr){3-5}\cmidrule(lr){6-8}\cmidrule(lr){9-11}
Model & Bits & U & Out & HG & U & Out & HG & U & Out & HG & BF16 \\
\midrule
LLaVA-1.5 & 2 & 0.303 & 0.387 & 0.375 & 0.443 & 0.437 & 0.427 & 0.460 & 0.442 & 0.427 & 0.453 \\
LLaVA-1.5 & 3 & 0.433 & 0.440 & 0.430 & 0.455 & 0.465 & 0.475 & 0.462 & 0.452 & 0.460 & 0.453 \\
LLaVA-1.5 & 4 & 0.440 & 0.440 & 0.438 & 0.448 & 0.453 & 0.443 & 0.433 & 0.455 & 0.432 & 0.453 \\
\addlinespace[2pt]
LLaVA-NeXT & 2 & 0.522 & 0.525 & 0.567 & 0.540 & 0.532 & 0.527 & 0.553 & 0.532 & 0.530 & 0.553 \\
LLaVA-NeXT & 3 & 0.553 & 0.532 & 0.525 & 0.558 & 0.542 & 0.542 & 0.558 & 0.562 & 0.542 & 0.553 \\
LLaVA-NeXT & 4 & 0.560 & 0.557 & 0.555 & 0.558 & 0.553 & 0.550 & 0.558 & 0.553 & 0.562 & 0.553 \\
\addlinespace[2pt]
Qwen2.5-VL & 2 & 0.007 & 0.647 & 0.715 & 0.038 & 0.758 & 0.795 & 0.738 & 0.727 & 0.758 & 0.770 \\
Qwen2.5-VL & 3 & 0.002 & 0.740 & 0.753 & 0.000 & 0.765 & 0.762 & 0.772 & 0.768 & 0.770 & 0.770 \\
Qwen2.5-VL & 4 & 0.017 & 0.762 & 0.758 & 0.062 & 0.788 & 0.775 & 0.760 & 0.767 & 0.770 & 0.770 \\
\addlinespace[2pt]
Qwen2-VL & 2 & 0.045 & 0.455 & 0.385 & 0.083 & 0.723 & 0.702 & 0.720 & 0.737 & 0.708 & 0.792 \\
Qwen2-VL & 3 & 0.013 & 0.757 & 0.762 & 0.210 & 0.752 & 0.772 & 0.765 & 0.777 & 0.758 & 0.792 \\
Qwen2-VL & 4 & 0.047 & 0.783 & 0.773 & 0.073 & 0.787 & 0.767 & 0.782 & 0.792 & 0.795 & 0.792 \\
\addlinespace[2pt]
Gemma-3 & 2 & 0.008 & 0.012 & 0.015 & 0.623 & 0.607 & 0.648 & 0.697 & 0.682 & 0.685 & 0.740 \\
Gemma-3 & 3 & 0.545 & 0.655 & 0.620 & 0.727 & 0.735 & 0.730 & 0.760 & 0.725 & 0.725 & 0.740 \\
Gemma-3 & 4 & 0.743 & 0.708 & 0.730 & 0.738 & 0.715 & 0.717 & 0.727 & 0.727 & 0.730 & 0.740 \\
\addlinespace[2pt]
Gemma-4 & 2 & 0.002 & 0.000 & 0.002 & 0.652 & 0.670 & 0.663 & 0.008 & 0.037 & 0.025 & 0.673 \\
Gemma-4 & 3 & 0.292 & 0.335 & 0.385 & 0.668 & 0.667 & 0.667 & 0.532 & 0.608 & 0.585 & 0.673 \\
Gemma-4 & 4 & 0.665 & 0.672 & 0.637 & 0.675 & 0.673 & 0.677 & 0.683 & 0.655 & 0.665 & 0.673 \\
\addlinespace[2pt]
Pixtral & 2 & 0.250 & 0.448 & 0.343 & 0.718 & 0.717 & 0.728 & 0.678 & 0.713 & 0.705 & 0.723 \\
Pixtral & 3 & 0.725 & 0.723 & 0.722 & 0.710 & 0.718 & 0.720 & 0.725 & 0.725 & 0.712 & 0.723 \\
Pixtral & 4 & 0.727 & 0.723 & 0.723 & 0.723 & 0.728 & 0.723 & 0.715 & 0.720 & 0.717 & 0.723 \\
\addlinespace[2pt]
InternVL3 & 2 & 0.028 & 0.512 & 0.485 & 0.093 & 0.637 & 0.568 & 0.627 & 0.645 & 0.662 & 0.673 \\
InternVL3 & 3 & 0.042 & 0.660 & 0.655 & 0.185 & 0.655 & 0.662 & 0.682 & 0.692 & 0.678 & 0.673 \\
InternVL3 & 4 & 0.040 & 0.693 & 0.680 & 0.400 & 0.673 & 0.658 & 0.682 & 0.678 & 0.693 & 0.673 \\
\midrule
Mean & 2 & 0.146 & 0.373 & 0.361 & 0.399 & 0.635 & 0.632 & 0.560 & 0.564 & 0.563 & 0.672 \\
Mean & 3 & 0.326 & 0.605 & 0.607 & 0.439 & 0.662 & 0.666 & 0.657 & 0.664 & 0.654 & 0.672 \\
Mean & 4 & 0.405 & 0.667 & 0.662 & 0.460 & 0.671 & 0.664 & 0.668 & 0.668 & 0.671 & 0.672 \\
\bottomrule
\end{tabular}
\end{table}

\clearpage
\begin{table}[!htbp]
\centering
\caption{\textbf{Capped captioning with POPE calibration.} ROUGE-L F1 against
BF16 output on $200$ examples, not human-caption quality. K+V protection
uses the one-eighth head budget. Mean averages all eight models.}
\label{tab:sweep_pope_caption}
\setlength{\tabcolsep}{4.25pt}
\renewcommand{\arraystretch}{1.12}
\begin{tabular}{@{}lc*{10}{c}@{}}
\toprule
 & & \multicolumn{3}{c}{min--max} & \multicolumn{3}{c}{AKVQ-style} & \multicolumn{3}{c}{KIVI-style} & \\
\cmidrule(lr){3-5}\cmidrule(lr){6-8}\cmidrule(lr){9-11}
Model & Bits & U & Out & HG & U & Out & HG & U & Out & HG & BF16 \\
\midrule
LLaVA-1.5 & 2 & 0.373 & 0.451 & 0.480 & 0.646 & 0.667 & 0.670 & 0.638 & 0.656 & 0.673 & 1.000 \\
LLaVA-1.5 & 3 & 0.738 & 0.745 & 0.748 & 0.788 & 0.797 & 0.814 & 0.791 & 0.802 & 0.804 & 1.000 \\
LLaVA-1.5 & 4 & 0.847 & 0.849 & 0.868 & 0.891 & 0.894 & 0.894 & 0.873 & 0.889 & 0.892 & 1.000 \\
\addlinespace[2pt]
LLaVA-NeXT & 2 & 0.460 & 0.505 & 0.509 & 0.626 & 0.661 & 0.660 & 0.551 & 0.600 & 0.602 & 1.000 \\
LLaVA-NeXT & 3 & 0.724 & 0.781 & 0.783 & 0.825 & 0.836 & 0.849 & 0.813 & 0.809 & 0.839 & 1.000 \\
LLaVA-NeXT & 4 & 0.885 & 0.852 & 0.887 & 0.902 & 0.897 & 0.882 & 0.881 & 0.867 & 0.867 & 1.000 \\
\addlinespace[2pt]
Qwen2.5-VL & 2 & 0.005 & 0.352 & 0.367 & 0.095 & 0.425 & 0.444 & 0.451 & 0.462 & 0.457 & 1.000 \\
Qwen2.5-VL & 3 & 0.009 & 0.426 & 0.430 & 0.022 & 0.453 & 0.444 & 0.602 & 0.606 & 0.614 & 1.000 \\
Qwen2.5-VL & 4 & 0.086 & 0.547 & 0.545 & 0.147 & 0.575 & 0.583 & 0.630 & 0.649 & 0.655 & 1.000 \\
\addlinespace[2pt]
Qwen2-VL & 2 & 0.053 & 0.289 & 0.292 & 0.015 & 0.416 & 0.437 & 0.380 & 0.383 & 0.396 & 1.000 \\
Qwen2-VL & 3 & 0.025 & 0.569 & 0.580 & 0.017 & 0.637 & 0.651 & 0.617 & 0.653 & 0.656 & 1.000 \\
Qwen2-VL & 4 & 0.008 & 0.734 & 0.751 & 0.336 & 0.725 & 0.713 & 0.751 & 0.761 & 0.743 & 1.000 \\
\addlinespace[2pt]
Gemma-3 & 2 & 0.073 & 0.156 & 0.131 & 0.518 & 0.545 & 0.542 & 0.580 & 0.601 & 0.601 & 1.000 \\
Gemma-3 & 3 & 0.532 & 0.570 & 0.570 & 0.675 & 0.719 & 0.719 & 0.705 & 0.735 & 0.750 & 1.000 \\
Gemma-3 & 4 & 0.681 & 0.696 & 0.700 & 0.766 & 0.801 & 0.822 & 0.798 & 0.836 & 0.822 & 1.000 \\
\addlinespace[2pt]
Gemma-4 & 2 & 0.037 & 0.052 & 0.049 & 0.667 & 0.657 & 0.676 & 0.096 & 0.133 & 0.132 & 1.000 \\
Gemma-4 & 3 & 0.393 & 0.394 & 0.397 & 0.762 & 0.758 & 0.763 & 0.471 & 0.470 & 0.481 & 1.000 \\
Gemma-4 & 4 & 0.600 & 0.616 & 0.628 & 0.834 & 0.837 & 0.835 & 0.613 & 0.636 & 0.649 & 1.000 \\
\addlinespace[2pt]
Pixtral & 2 & 0.262 & 0.301 & 0.319 & 0.616 & 0.677 & 0.696 & 0.578 & 0.583 & 0.632 & 1.000 \\
Pixtral & 3 & 0.763 & 0.789 & 0.783 & 0.820 & 0.853 & 0.864 & 0.810 & 0.816 & 0.799 & 1.000 \\
Pixtral & 4 & 0.853 & 0.883 & 0.889 & 0.892 & 0.910 & 0.899 & 0.869 & 0.893 & 0.903 & 1.000 \\
\addlinespace[2pt]
InternVL3 & 2 & 0.071 & 0.343 & 0.349 & 0.099 & 0.467 & 0.403 & 0.491 & 0.502 & 0.506 & 1.000 \\
InternVL3 & 3 & 0.045 & 0.468 & 0.463 & 0.104 & 0.568 & 0.505 & 0.672 & 0.685 & 0.694 & 1.000 \\
InternVL3 & 4 & 0.134 & 0.586 & 0.532 & 0.279 & 0.570 & 0.528 & 0.736 & 0.740 & 0.749 & 1.000 \\
\midrule
Mean & 2 & 0.167 & 0.306 & 0.312 & 0.410 & 0.564 & 0.566 & 0.471 & 0.490 & 0.500 & 1.000 \\
Mean & 3 & 0.404 & 0.593 & 0.594 & 0.502 & 0.703 & 0.701 & 0.685 & 0.697 & 0.705 & 1.000 \\
Mean & 4 & 0.512 & 0.720 & 0.725 & 0.631 & 0.776 & 0.770 & 0.769 & 0.784 & 0.785 & 1.000 \\
\bottomrule
\end{tabular}
\end{table}

\clearpage
\begin{table}[!htbp]
\centering
\caption{\textbf{Capped TextVQA with MMBench calibration.} Same $200$-question
scope, metric, head budget, and columns as Table~\ref{tab:sweep_pope_textvqa}.
Mean averages all eight models.}
\label{tab:sweep_mmbench_textvqa}
\setlength{\tabcolsep}{4.25pt}
\renewcommand{\arraystretch}{1.12}
\begin{tabular}{@{}lc*{10}{c}@{}}
\toprule
 & & \multicolumn{3}{c}{min--max} & \multicolumn{3}{c}{AKVQ-style} & \multicolumn{3}{c}{KIVI-style} & \\
\cmidrule(lr){3-5}\cmidrule(lr){6-8}\cmidrule(lr){9-11}
Model & Bits & U & Out & HG & U & Out & HG & U & Out & HG & BF16 \\
\midrule
LLaVA-1.5 & 2 & 0.303 & 0.367 & 0.368 & 0.443 & 0.432 & 0.410 & 0.460 & 0.437 & 0.428 & 0.453 \\
LLaVA-1.5 & 3 & 0.433 & 0.433 & 0.450 & 0.455 & 0.455 & 0.467 & 0.462 & 0.470 & 0.478 & 0.453 \\
LLaVA-1.5 & 4 & 0.440 & 0.432 & 0.440 & 0.448 & 0.457 & 0.453 & 0.433 & 0.432 & 0.445 & 0.453 \\
\addlinespace[2pt]
LLaVA-NeXT & 2 & 0.522 & 0.507 & 0.548 & 0.540 & 0.533 & 0.528 & 0.553 & 0.525 & 0.548 & 0.553 \\
LLaVA-NeXT & 3 & 0.553 & 0.558 & 0.537 & 0.558 & 0.560 & 0.547 & 0.558 & 0.572 & 0.548 & 0.553 \\
LLaVA-NeXT & 4 & 0.560 & 0.570 & 0.552 & 0.558 & 0.543 & 0.543 & 0.558 & 0.560 & 0.560 & 0.553 \\
\addlinespace[2pt]
Qwen2.5-VL & 2 & 0.007 & 0.660 & 0.347 & 0.038 & 0.748 & 0.380 & 0.738 & 0.710 & 0.740 & 0.770 \\
Qwen2.5-VL & 3 & 0.002 & 0.770 & 0.538 & 0.000 & 0.768 & 0.598 & 0.772 & 0.745 & 0.773 & 0.770 \\
Qwen2.5-VL & 4 & 0.017 & 0.763 & 0.620 & 0.062 & 0.792 & 0.688 & 0.760 & 0.777 & 0.777 & 0.770 \\
\addlinespace[2pt]
Qwen2-VL & 2 & 0.045 & 0.452 & 0.288 & 0.083 & 0.685 & 0.460 & 0.720 & 0.730 & 0.723 & 0.792 \\
Qwen2-VL & 3 & 0.013 & 0.757 & 0.762 & 0.210 & 0.772 & 0.748 & 0.765 & 0.753 & 0.785 & 0.792 \\
Qwen2-VL & 4 & 0.047 & 0.785 & 0.577 & 0.073 & 0.772 & 0.758 & 0.782 & 0.792 & 0.785 & 0.792 \\
\addlinespace[2pt]
Gemma-3 & 2 & 0.008 & 0.035 & 0.010 & 0.623 & 0.650 & 0.632 & 0.697 & 0.702 & 0.690 & 0.740 \\
Gemma-3 & 3 & 0.545 & 0.577 & 0.593 & 0.727 & 0.748 & 0.730 & 0.760 & 0.728 & 0.733 & 0.740 \\
Gemma-3 & 4 & 0.743 & 0.718 & 0.712 & 0.738 & 0.723 & 0.725 & 0.727 & 0.725 & 0.743 & 0.740 \\
\addlinespace[2pt]
Gemma-4 & 2 & 0.002 & 0.003 & 0.000 & 0.652 & 0.648 & 0.632 & 0.008 & 0.018 & 0.035 & 0.673 \\
Gemma-4 & 3 & 0.292 & 0.293 & 0.372 & 0.668 & 0.657 & 0.663 & 0.532 & 0.548 & 0.550 & 0.673 \\
Gemma-4 & 4 & 0.665 & 0.673 & 0.670 & 0.675 & 0.678 & 0.673 & 0.683 & 0.663 & 0.663 & 0.673 \\
\addlinespace[2pt]
Pixtral & 2 & 0.250 & 0.413 & 0.407 & 0.718 & 0.718 & 0.713 & 0.678 & 0.688 & 0.678 & 0.723 \\
Pixtral & 3 & 0.725 & 0.730 & 0.730 & 0.710 & 0.733 & 0.710 & 0.725 & 0.717 & 0.730 & 0.723 \\
Pixtral & 4 & 0.727 & 0.725 & 0.723 & 0.723 & 0.723 & 0.723 & 0.715 & 0.723 & 0.728 & 0.723 \\
\addlinespace[2pt]
InternVL3 & 2 & 0.028 & 0.500 & 0.462 & 0.093 & 0.642 & 0.587 & 0.627 & 0.663 & 0.637 & 0.673 \\
InternVL3 & 3 & 0.042 & 0.683 & 0.695 & 0.185 & 0.682 & 0.638 & 0.682 & 0.672 & 0.672 & 0.673 \\
InternVL3 & 4 & 0.040 & 0.670 & 0.663 & 0.400 & 0.675 & 0.642 & 0.682 & 0.678 & 0.688 & 0.673 \\
\midrule
Mean & 2 & 0.146 & 0.367 & 0.304 & 0.399 & 0.632 & 0.543 & 0.560 & 0.559 & 0.560 & 0.672 \\
Mean & 3 & 0.326 & 0.600 & 0.585 & 0.439 & 0.672 & 0.638 & 0.657 & 0.651 & 0.659 & 0.672 \\
Mean & 4 & 0.405 & 0.667 & 0.620 & 0.460 & 0.670 & 0.651 & 0.668 & 0.669 & 0.674 & 0.672 \\
\bottomrule
\end{tabular}
\end{table}

\clearpage
\begin{table}[!htbp]
\centering
\caption{\textbf{Capped captioning with MMBench calibration.} Same $200$-example
scope, BF16-output ROUGE-L metric, head budget, and columns as
Table~\ref{tab:sweep_pope_caption}. Mean averages all eight models.}
\label{tab:sweep_mmbench_caption}
\setlength{\tabcolsep}{4.25pt}
\renewcommand{\arraystretch}{1.12}
\begin{tabular}{@{}lc*{10}{c}@{}}
\toprule
 & & \multicolumn{3}{c}{min--max} & \multicolumn{3}{c}{AKVQ-style} & \multicolumn{3}{c}{KIVI-style} & \\
\cmidrule(lr){3-5}\cmidrule(lr){6-8}\cmidrule(lr){9-11}
Model & Bits & U & Out & HG & U & Out & HG & U & Out & HG & BF16 \\
\midrule
LLaVA-1.5 & 2 & 0.373 & 0.482 & 0.507 & 0.646 & 0.649 & 0.694 & 0.638 & 0.669 & 0.657 & 1.000 \\
LLaVA-1.5 & 3 & 0.738 & 0.739 & 0.753 & 0.788 & 0.811 & 0.824 & 0.791 & 0.797 & 0.819 & 1.000 \\
LLaVA-1.5 & 4 & 0.847 & 0.857 & 0.841 & 0.891 & 0.880 & 0.892 & 0.873 & 0.874 & 0.864 & 1.000 \\
\addlinespace[2pt]
LLaVA-NeXT & 2 & 0.460 & 0.495 & 0.514 & 0.626 & 0.643 & 0.661 & 0.551 & 0.597 & 0.593 & 1.000 \\
LLaVA-NeXT & 3 & 0.724 & 0.757 & 0.765 & 0.825 & 0.836 & 0.861 & 0.813 & 0.803 & 0.843 & 1.000 \\
LLaVA-NeXT & 4 & 0.885 & 0.863 & 0.870 & 0.902 & 0.887 & 0.892 & 0.881 & 0.871 & 0.872 & 1.000 \\
\addlinespace[2pt]
Qwen2.5-VL & 2 & 0.005 & 0.353 & 0.292 & 0.095 & 0.422 & 0.299 & 0.451 & 0.461 & 0.481 & 1.000 \\
Qwen2.5-VL & 3 & 0.009 & 0.472 & 0.348 & 0.022 & 0.562 & 0.364 & 0.602 & 0.606 & 0.591 & 1.000 \\
Qwen2.5-VL & 4 & 0.086 & 0.609 & 0.400 & 0.147 & 0.616 & 0.415 & 0.630 & 0.655 & 0.676 & 1.000 \\
\addlinespace[2pt]
Qwen2-VL & 2 & 0.053 & 0.296 & 0.215 & 0.015 & 0.420 & 0.160 & 0.380 & 0.382 & 0.402 & 1.000 \\
Qwen2-VL & 3 & 0.025 & 0.583 & 0.153 & 0.017 & 0.633 & 0.497 & 0.617 & 0.621 & 0.649 & 1.000 \\
Qwen2-VL & 4 & 0.008 & 0.743 & 0.191 & 0.336 & 0.722 & 0.600 & 0.751 & 0.756 & 0.780 & 1.000 \\
\addlinespace[2pt]
Gemma-3 & 2 & 0.073 & 0.166 & 0.132 & 0.518 & 0.547 & 0.544 & 0.580 & 0.599 & 0.611 & 1.000 \\
Gemma-3 & 3 & 0.532 & 0.555 & 0.561 & 0.675 & 0.725 & 0.702 & 0.705 & 0.738 & 0.749 & 1.000 \\
Gemma-3 & 4 & 0.681 & 0.691 & 0.692 & 0.766 & 0.801 & 0.797 & 0.798 & 0.800 & 0.821 & 1.000 \\
\addlinespace[2pt]
Gemma-4 & 2 & 0.037 & 0.038 & 0.055 & 0.667 & 0.658 & 0.664 & 0.096 & 0.147 & 0.118 & 1.000 \\
Gemma-4 & 3 & 0.393 & 0.398 & 0.411 & 0.762 & 0.773 & 0.778 & 0.471 & 0.489 & 0.483 & 1.000 \\
Gemma-4 & 4 & 0.600 & 0.619 & 0.618 & 0.834 & 0.826 & 0.807 & 0.613 & 0.636 & 0.661 & 1.000 \\
\addlinespace[2pt]
Pixtral & 2 & 0.262 & 0.315 & 0.307 & 0.616 & 0.691 & 0.698 & 0.578 & 0.580 & 0.576 & 1.000 \\
Pixtral & 3 & 0.763 & 0.767 & 0.786 & 0.820 & 0.838 & 0.836 & 0.810 & 0.819 & 0.808 & 1.000 \\
Pixtral & 4 & 0.853 & 0.901 & 0.898 & 0.892 & 0.909 & 0.922 & 0.869 & 0.890 & 0.894 & 1.000 \\
\addlinespace[2pt]
InternVL3 & 2 & 0.071 & 0.378 & 0.356 & 0.099 & 0.489 & 0.408 & 0.491 & 0.505 & 0.501 & 1.000 \\
InternVL3 & 3 & 0.045 & 0.553 & 0.455 & 0.104 & 0.665 & 0.509 & 0.672 & 0.685 & 0.681 & 1.000 \\
InternVL3 & 4 & 0.134 & 0.702 & 0.534 & 0.279 & 0.727 & 0.530 & 0.736 & 0.755 & 0.755 & 1.000 \\
\midrule
Mean & 2 & 0.167 & 0.315 & 0.297 & 0.410 & 0.565 & 0.516 & 0.471 & 0.493 & 0.492 & 1.000 \\
Mean & 3 & 0.404 & 0.603 & 0.529 & 0.502 & 0.730 & 0.671 & 0.685 & 0.695 & 0.703 & 1.000 \\
Mean & 4 & 0.512 & 0.748 & 0.631 & 0.631 & 0.796 & 0.732 & 0.769 & 0.780 & 0.790 & 1.000 \\
\bottomrule
\end{tabular}
\end{table}

\clearpage
\section{Additional Capped Head Budgets}
\label{app:additional_budgets}
The following tables extend the capped results to protected-head fractions
$1/16$, $1/12$, and $1/4$ under POPE and MMBench calibration.
Six-task accuracy, TextVQA, and caption fidelity are reported separately;
each Mean uses all eight models within its metric, base, precision, and budget.
The fractions are nominal: the protected head count is rounded for each model.
For six-task evaluation, the calibration benchmark uses $900$ held-out questions,
MMMU $839$, and the other tasks $1000$, consistently across bases.
Throughout these tables, U means no head protection, Out means output-only
selection, and HG means HeadGuard (image-sensitivity plus output-sensitivity scores).
Entries are accuracies or task scores, not gains.
\begin{table}[!htbp]
\centering
\caption{\textbf{Six-task accuracy, $1/16$ budget, POPE calibration.} K+V protection with $K=\operatorname{round}(N_{\rm KV}/16)$. Equal-weight accuracy over POPE, MME, MMBench, ScienceQA, MMMU, and SEED; up to $1000$ questions/task ($900$ for POPE; $839$ for MMMU). U = uniform; Out = output-only; HG = hybrid. Mean averages all eight models.}
\label{tab:budget_d16_pope_six}
\setlength{\tabcolsep}{4.25pt}
\renewcommand{\arraystretch}{1.12}
\begin{tabular}{@{}lc*{10}{c}@{}}
\toprule
 & & \multicolumn{3}{c}{min--max} & \multicolumn{3}{c}{AKVQ-style} & \multicolumn{3}{c}{KIVI-style} & \\
\cmidrule(lr){3-5}\cmidrule(lr){6-8}\cmidrule(lr){9-11}
Model & Bits & U & Out & HG & U & Out & HG & U & Out & HG & BF16 \\
\midrule
LLaVA-1.5 & 2 & 0.515 & 0.542 & 0.543 & 0.637 & 0.635 & 0.641 & 0.638 & 0.633 & 0.637 & 0.643 \\
LLaVA-1.5 & 3 & 0.640 & 0.641 & 0.641 & 0.642 & 0.640 & 0.643 & 0.644 & 0.644 & 0.647 & 0.643 \\
LLaVA-1.5 & 4 & 0.643 & 0.644 & 0.644 & 0.644 & 0.642 & 0.642 & 0.642 & 0.642 & 0.645 & 0.643 \\
\addlinespace[2pt]
LLaVA-NeXT & 2 & 0.618 & 0.644 & 0.649 & 0.663 & 0.668 & 0.662 & 0.656 & 0.661 & 0.661 & 0.666 \\
LLaVA-NeXT & 3 & 0.667 & 0.670 & 0.669 & 0.668 & 0.666 & 0.665 & 0.666 & 0.667 & 0.665 & 0.666 \\
LLaVA-NeXT & 4 & 0.667 & 0.667 & 0.667 & 0.665 & 0.666 & 0.665 & 0.667 & 0.665 & 0.665 & 0.666 \\
\bottomrule
\end{tabular}
\end{table}

\clearpage
\begin{table}[!htbp]
\centering
\caption{\textbf{TextVQA, $1/16$ budget, POPE calibration.} K+V protection with $K=\operatorname{round}(N_{\rm KV}/16)$. Custom VQA-style score on $200$ questions. U = uniform; Out = output-only; HG = hybrid. Mean averages all eight models.}
\label{tab:budget_d16_pope_textvqa}
\setlength{\tabcolsep}{4.25pt}
\renewcommand{\arraystretch}{1.12}
\begin{tabular}{@{}lc*{10}{c}@{}}
\toprule
 & & \multicolumn{3}{c}{min--max} & \multicolumn{3}{c}{AKVQ-style} & \multicolumn{3}{c}{KIVI-style} & \\
\cmidrule(lr){3-5}\cmidrule(lr){6-8}\cmidrule(lr){9-11}
Model & Bits & U & Out & HG & U & Out & HG & U & Out & HG & BF16 \\
\midrule
LLaVA-1.5 & 2 & 0.303 & 0.350 & 0.313 & 0.443 & 0.440 & 0.447 & 0.460 & 0.437 & 0.428 & 0.453 \\
LLaVA-1.5 & 3 & 0.433 & 0.443 & 0.435 & 0.455 & 0.450 & 0.458 & 0.462 & 0.452 & 0.463 & 0.453 \\
LLaVA-1.5 & 4 & 0.440 & 0.425 & 0.428 & 0.448 & 0.458 & 0.445 & 0.433 & 0.442 & 0.440 & 0.453 \\
\addlinespace[2pt]
LLaVA-NeXT & 2 & 0.522 & 0.515 & 0.532 & 0.540 & 0.558 & 0.517 & 0.553 & 0.533 & 0.535 & 0.553 \\
LLaVA-NeXT & 3 & 0.553 & 0.548 & 0.545 & 0.558 & 0.550 & 0.545 & 0.558 & 0.553 & 0.558 & 0.553 \\
LLaVA-NeXT & 4 & 0.560 & 0.560 & 0.547 & 0.558 & 0.548 & 0.552 & 0.558 & 0.548 & 0.557 & 0.553 \\
\bottomrule
\end{tabular}
\end{table}

\clearpage
\begin{table}[!htbp]
\centering
\caption{\textbf{Caption fidelity, $1/16$ budget, POPE calibration.} K+V protection with $K=\operatorname{round}(N_{\rm KV}/16)$. ROUGE-L F1 against BF16 output on $200$ examples, not human-caption quality. U = uniform; Out = output-only; HG = hybrid. Mean averages all eight models.}
\label{tab:budget_d16_pope_caption}
\setlength{\tabcolsep}{4.25pt}
\renewcommand{\arraystretch}{1.12}
\begin{tabular}{@{}lc*{10}{c}@{}}
\toprule
 & & \multicolumn{3}{c}{min--max} & \multicolumn{3}{c}{AKVQ-style} & \multicolumn{3}{c}{KIVI-style} & \\
\cmidrule(lr){3-5}\cmidrule(lr){6-8}\cmidrule(lr){9-11}
Model & Bits & U & Out & HG & U & Out & HG & U & Out & HG & BF16 \\
\midrule
LLaVA-1.5 & 2 & 0.373 & 0.445 & 0.415 & 0.646 & 0.647 & 0.641 & 0.638 & 0.648 & 0.662 & 1.000 \\
LLaVA-1.5 & 3 & 0.738 & 0.740 & 0.754 & 0.788 & 0.783 & 0.807 & 0.791 & 0.804 & 0.816 & 1.000 \\
LLaVA-1.5 & 4 & 0.847 & 0.858 & 0.852 & 0.891 & 0.886 & 0.879 & 0.873 & 0.868 & 0.868 & 1.000 \\
\addlinespace[2pt]
LLaVA-NeXT & 2 & 0.460 & 0.494 & 0.500 & 0.626 & 0.650 & 0.661 & 0.551 & 0.566 & 0.596 & 1.000 \\
LLaVA-NeXT & 3 & 0.724 & 0.745 & 0.770 & 0.825 & 0.838 & 0.848 & 0.813 & 0.791 & 0.822 & 1.000 \\
LLaVA-NeXT & 4 & 0.885 & 0.873 & 0.863 & 0.902 & 0.906 & 0.894 & 0.881 & 0.853 & 0.886 & 1.000 \\
\bottomrule
\end{tabular}
\end{table}

\clearpage
\begin{table}[!htbp]
\centering
\caption{\textbf{Six-task accuracy, $1/16$ budget, MMBench calibration.} K+V protection with $K=\operatorname{round}(N_{\rm KV}/16)$. Equal-weight accuracy over POPE, MME, MMBench, ScienceQA, MMMU, and SEED; up to $1000$ questions/task ($900$ for MMBench; $839$ for MMMU). U = uniform; Out = output-only; HG = hybrid. Mean averages all eight models.}
\label{tab:budget_d16_mmbench_six}
\setlength{\tabcolsep}{4.25pt}
\renewcommand{\arraystretch}{1.12}
\begin{tabular}{@{}lc*{10}{c}@{}}
\toprule
 & & \multicolumn{3}{c}{min--max} & \multicolumn{3}{c}{AKVQ-style} & \multicolumn{3}{c}{KIVI-style} & \\
\cmidrule(lr){3-5}\cmidrule(lr){6-8}\cmidrule(lr){9-11}
Model & Bits & U & Out & HG & U & Out & HG & U & Out & HG & BF16 \\
\midrule
LLaVA-1.5 & 2 & 0.514 & 0.561 & 0.541 & 0.639 & 0.639 & 0.638 & 0.639 & 0.636 & 0.643 & 0.644 \\
LLaVA-1.5 & 3 & 0.641 & 0.639 & 0.643 & 0.643 & 0.644 & 0.642 & 0.645 & 0.645 & 0.644 & 0.644 \\
LLaVA-1.5 & 4 & 0.644 & 0.644 & 0.644 & 0.646 & 0.645 & 0.646 & 0.643 & 0.645 & 0.645 & 0.644 \\
\addlinespace[2pt]
LLaVA-NeXT & 2 & 0.620 & 0.626 & 0.645 & 0.663 & 0.666 & 0.665 & 0.656 & 0.659 & 0.662 & 0.666 \\
LLaVA-NeXT & 3 & 0.666 & 0.668 & 0.668 & 0.668 & 0.668 & 0.666 & 0.667 & 0.667 & 0.666 & 0.666 \\
LLaVA-NeXT & 4 & 0.668 & 0.666 & 0.666 & 0.666 & 0.669 & 0.669 & 0.668 & 0.665 & 0.667 & 0.666 \\
\bottomrule
\end{tabular}
\end{table}

\clearpage
\begin{table}[!htbp]
\centering
\caption{\textbf{TextVQA, $1/16$ budget, MMBench calibration.} K+V protection with $K=\operatorname{round}(N_{\rm KV}/16)$. Custom VQA-style score on $200$ questions. U = uniform; Out = output-only; HG = hybrid. Mean averages all eight models.}
\label{tab:budget_d16_mmbench_textvqa}
\setlength{\tabcolsep}{4.25pt}
\renewcommand{\arraystretch}{1.12}
\begin{tabular}{@{}lc*{10}{c}@{}}
\toprule
 & & \multicolumn{3}{c}{min--max} & \multicolumn{3}{c}{AKVQ-style} & \multicolumn{3}{c}{KIVI-style} & \\
\cmidrule(lr){3-5}\cmidrule(lr){6-8}\cmidrule(lr){9-11}
Model & Bits & U & Out & HG & U & Out & HG & U & Out & HG & BF16 \\
\midrule
LLaVA-1.5 & 2 & 0.303 & 0.332 & 0.325 & 0.443 & 0.428 & 0.437 & 0.460 & 0.430 & 0.447 & 0.453 \\
LLaVA-1.5 & 3 & 0.433 & 0.438 & 0.432 & 0.455 & 0.460 & 0.468 & 0.462 & 0.452 & 0.467 & 0.453 \\
LLaVA-1.5 & 4 & 0.440 & 0.432 & 0.435 & 0.448 & 0.448 & 0.445 & 0.433 & 0.442 & 0.450 & 0.453 \\
\addlinespace[2pt]
LLaVA-NeXT & 2 & 0.522 & 0.483 & 0.512 & 0.540 & 0.515 & 0.538 & 0.553 & 0.547 & 0.532 & 0.553 \\
LLaVA-NeXT & 3 & 0.553 & 0.552 & 0.533 & 0.558 & 0.560 & 0.555 & 0.558 & 0.533 & 0.550 & 0.553 \\
LLaVA-NeXT & 4 & 0.560 & 0.565 & 0.560 & 0.558 & 0.558 & 0.548 & 0.558 & 0.557 & 0.558 & 0.553 \\
\bottomrule
\end{tabular}
\end{table}

\clearpage
\begin{table}[!htbp]
\centering
\caption{\textbf{Caption fidelity, $1/16$ budget, MMBench calibration.} K+V protection with $K=\operatorname{round}(N_{\rm KV}/16)$. ROUGE-L F1 against BF16 output on $200$ examples, not human-caption quality. U = uniform; Out = output-only; HG = hybrid. Mean averages all eight models.}
\label{tab:budget_d16_mmbench_caption}
\setlength{\tabcolsep}{4.25pt}
\renewcommand{\arraystretch}{1.12}
\begin{tabular}{@{}lc*{10}{c}@{}}
\toprule
 & & \multicolumn{3}{c}{min--max} & \multicolumn{3}{c}{AKVQ-style} & \multicolumn{3}{c}{KIVI-style} & \\
\cmidrule(lr){3-5}\cmidrule(lr){6-8}\cmidrule(lr){9-11}
Model & Bits & U & Out & HG & U & Out & HG & U & Out & HG & BF16 \\
\midrule
LLaVA-1.5 & 2 & 0.373 & 0.457 & 0.430 & 0.646 & 0.647 & 0.659 & 0.638 & 0.642 & 0.653 & 1.000 \\
LLaVA-1.5 & 3 & 0.738 & 0.735 & 0.748 & 0.788 & 0.813 & 0.815 & 0.791 & 0.801 & 0.796 & 1.000 \\
LLaVA-1.5 & 4 & 0.847 & 0.836 & 0.838 & 0.891 & 0.874 & 0.892 & 0.873 & 0.862 & 0.888 & 1.000 \\
\addlinespace[2pt]
LLaVA-NeXT & 2 & 0.460 & 0.493 & 0.495 & 0.626 & 0.637 & 0.667 & 0.551 & 0.571 & 0.584 & 1.000 \\
LLaVA-NeXT & 3 & 0.724 & 0.753 & 0.743 & 0.825 & 0.853 & 0.844 & 0.813 & 0.800 & 0.812 & 1.000 \\
LLaVA-NeXT & 4 & 0.885 & 0.860 & 0.866 & 0.902 & 0.888 & 0.882 & 0.881 & 0.878 & 0.857 & 1.000 \\
\bottomrule
\end{tabular}
\end{table}

\clearpage
\begin{table}[!htbp]
\centering
\caption{\textbf{Six-task accuracy, $1/12$ budget, POPE calibration.} K+V protection with $K=\operatorname{round}(N_{\rm KV}/12)$. Equal-weight accuracy over POPE, MME, MMBench, ScienceQA, MMMU, and SEED; up to $1000$ questions/task ($900$ for POPE; $839$ for MMMU). U = uniform; Out = output-only; HG = hybrid. Mean averages all eight models.}
\label{tab:budget_d12_pope_six}
\setlength{\tabcolsep}{4.25pt}
\renewcommand{\arraystretch}{1.12}
\begin{tabular}{@{}lc*{10}{c}@{}}
\toprule
 & & \multicolumn{3}{c}{min--max} & \multicolumn{3}{c}{AKVQ-style} & \multicolumn{3}{c}{KIVI-style} & \\
\cmidrule(lr){3-5}\cmidrule(lr){6-8}\cmidrule(lr){9-11}
Model & Bits & U & Out & HG & U & Out & HG & U & Out & HG & BF16 \\
\midrule
LLaVA-1.5 & 2 & 0.515 & 0.549 & 0.566 & 0.637 & 0.631 & 0.640 & 0.638 & 0.634 & 0.641 & 0.643 \\
LLaVA-1.5 & 3 & 0.640 & 0.642 & 0.645 & 0.642 & 0.641 & 0.642 & 0.644 & 0.643 & 0.643 & 0.643 \\
LLaVA-1.5 & 4 & 0.643 & 0.642 & 0.645 & 0.644 & 0.642 & 0.644 & 0.642 & 0.643 & 0.643 & 0.643 \\
\addlinespace[2pt]
LLaVA-NeXT & 2 & 0.618 & 0.644 & 0.653 & 0.663 & 0.666 & 0.668 & 0.656 & 0.661 & 0.666 & 0.666 \\
LLaVA-NeXT & 3 & 0.667 & 0.671 & 0.671 & 0.668 & 0.667 & 0.665 & 0.666 & 0.666 & 0.666 & 0.666 \\
LLaVA-NeXT & 4 & 0.667 & 0.666 & 0.665 & 0.665 & 0.667 & 0.667 & 0.667 & 0.667 & 0.665 & 0.666 \\
\bottomrule
\end{tabular}
\end{table}

\clearpage
\begin{table}[!htbp]
\centering
\caption{\textbf{TextVQA, $1/12$ budget, POPE calibration.} K+V protection with $K=\operatorname{round}(N_{\rm KV}/12)$. Custom VQA-style score on $200$ questions. U = uniform; Out = output-only; HG = hybrid. Mean averages all eight models.}
\label{tab:budget_d12_pope_textvqa}
\setlength{\tabcolsep}{4.25pt}
\renewcommand{\arraystretch}{1.12}
\begin{tabular}{@{}lc*{10}{c}@{}}
\toprule
 & & \multicolumn{3}{c}{min--max} & \multicolumn{3}{c}{AKVQ-style} & \multicolumn{3}{c}{KIVI-style} & \\
\cmidrule(lr){3-5}\cmidrule(lr){6-8}\cmidrule(lr){9-11}
Model & Bits & U & Out & HG & U & Out & HG & U & Out & HG & BF16 \\
\midrule
LLaVA-1.5 & 2 & 0.303 & 0.352 & 0.362 & 0.443 & 0.443 & 0.440 & 0.460 & 0.440 & 0.440 & 0.453 \\
LLaVA-1.5 & 3 & 0.433 & 0.423 & 0.430 & 0.455 & 0.452 & 0.442 & 0.462 & 0.438 & 0.465 & 0.453 \\
LLaVA-1.5 & 4 & 0.440 & 0.433 & 0.440 & 0.448 & 0.450 & 0.448 & 0.433 & 0.445 & 0.435 & 0.453 \\
\addlinespace[2pt]
LLaVA-NeXT & 2 & 0.522 & 0.545 & 0.522 & 0.540 & 0.525 & 0.550 & 0.553 & 0.555 & 0.543 & 0.553 \\
LLaVA-NeXT & 3 & 0.553 & 0.542 & 0.523 & 0.558 & 0.555 & 0.533 & 0.558 & 0.538 & 0.553 & 0.553 \\
LLaVA-NeXT & 4 & 0.560 & 0.555 & 0.555 & 0.558 & 0.558 & 0.555 & 0.558 & 0.547 & 0.550 & 0.553 \\
\bottomrule
\end{tabular}
\end{table}

\clearpage
\begin{table}[!htbp]
\centering
\caption{\textbf{Caption fidelity, $1/12$ budget, POPE calibration.} K+V protection with $K=\operatorname{round}(N_{\rm KV}/12)$. ROUGE-L F1 against BF16 output on $200$ examples, not human-caption quality. U = uniform; Out = output-only; HG = hybrid. Mean averages all eight models.}
\label{tab:budget_d12_pope_caption}
\setlength{\tabcolsep}{4.25pt}
\renewcommand{\arraystretch}{1.12}
\begin{tabular}{@{}lc*{10}{c}@{}}
\toprule
 & & \multicolumn{3}{c}{min--max} & \multicolumn{3}{c}{AKVQ-style} & \multicolumn{3}{c}{KIVI-style} & \\
\cmidrule(lr){3-5}\cmidrule(lr){6-8}\cmidrule(lr){9-11}
Model & Bits & U & Out & HG & U & Out & HG & U & Out & HG & BF16 \\
\midrule
LLaVA-1.5 & 2 & 0.373 & 0.446 & 0.458 & 0.646 & 0.646 & 0.652 & 0.638 & 0.641 & 0.644 & 1.000 \\
LLaVA-1.5 & 3 & 0.738 & 0.744 & 0.737 & 0.788 & 0.816 & 0.813 & 0.791 & 0.795 & 0.807 & 1.000 \\
LLaVA-1.5 & 4 & 0.847 & 0.841 & 0.848 & 0.891 & 0.886 & 0.884 & 0.873 & 0.876 & 0.866 & 1.000 \\
\addlinespace[2pt]
LLaVA-NeXT & 2 & 0.460 & 0.506 & 0.493 & 0.626 & 0.632 & 0.646 & 0.551 & 0.585 & 0.601 & 1.000 \\
LLaVA-NeXT & 3 & 0.724 & 0.760 & 0.778 & 0.825 & 0.852 & 0.850 & 0.813 & 0.806 & 0.802 & 1.000 \\
LLaVA-NeXT & 4 & 0.885 & 0.873 & 0.872 & 0.902 & 0.890 & 0.905 & 0.881 & 0.875 & 0.860 & 1.000 \\
\bottomrule
\end{tabular}
\end{table}

\clearpage
\begin{table}[!htbp]
\centering
\caption{\textbf{Six-task accuracy, $1/12$ budget, MMBench calibration.} K+V protection with $K=\operatorname{round}(N_{\rm KV}/12)$. Equal-weight accuracy over POPE, MME, MMBench, ScienceQA, MMMU, and SEED; up to $1000$ questions/task ($900$ for MMBench; $839$ for MMMU). U = uniform; Out = output-only; HG = hybrid. Mean averages all eight models.}
\label{tab:budget_d12_mmbench_six}
\setlength{\tabcolsep}{4.25pt}
\renewcommand{\arraystretch}{1.12}
\begin{tabular}{@{}lc*{10}{c}@{}}
\toprule
 & & \multicolumn{3}{c}{min--max} & \multicolumn{3}{c}{AKVQ-style} & \multicolumn{3}{c}{KIVI-style} & \\
\cmidrule(lr){3-5}\cmidrule(lr){6-8}\cmidrule(lr){9-11}
Model & Bits & U & Out & HG & U & Out & HG & U & Out & HG & BF16 \\
\midrule
LLaVA-1.5 & 2 & 0.514 & 0.560 & 0.547 & 0.639 & 0.637 & 0.640 & 0.639 & 0.635 & 0.640 & 0.644 \\
LLaVA-1.5 & 3 & 0.641 & 0.644 & 0.643 & 0.643 & 0.645 & 0.644 & 0.645 & 0.646 & 0.645 & 0.644 \\
LLaVA-1.5 & 4 & 0.644 & 0.644 & 0.646 & 0.646 & 0.647 & 0.646 & 0.643 & 0.646 & 0.645 & 0.644 \\
\addlinespace[2pt]
LLaVA-NeXT & 2 & 0.620 & 0.622 & 0.649 & 0.663 & 0.670 & 0.662 & 0.656 & 0.659 & 0.664 & 0.666 \\
LLaVA-NeXT & 3 & 0.666 & 0.670 & 0.669 & 0.668 & 0.666 & 0.664 & 0.667 & 0.667 & 0.665 & 0.666 \\
LLaVA-NeXT & 4 & 0.668 & 0.665 & 0.667 & 0.666 & 0.668 & 0.667 & 0.668 & 0.666 & 0.665 & 0.666 \\
\bottomrule
\end{tabular}
\end{table}

\clearpage
\begin{table}[!htbp]
\centering
\caption{\textbf{TextVQA, $1/12$ budget, MMBench calibration.} K+V protection with $K=\operatorname{round}(N_{\rm KV}/12)$. Custom VQA-style score on $200$ questions. U = uniform; Out = output-only; HG = hybrid. Mean averages all eight models.}
\label{tab:budget_d12_mmbench_textvqa}
\setlength{\tabcolsep}{4.25pt}
\renewcommand{\arraystretch}{1.12}
\begin{tabular}{@{}lc*{10}{c}@{}}
\toprule
 & & \multicolumn{3}{c}{min--max} & \multicolumn{3}{c}{AKVQ-style} & \multicolumn{3}{c}{KIVI-style} & \\
\cmidrule(lr){3-5}\cmidrule(lr){6-8}\cmidrule(lr){9-11}
Model & Bits & U & Out & HG & U & Out & HG & U & Out & HG & BF16 \\
\midrule
LLaVA-1.5 & 2 & 0.303 & 0.355 & 0.308 & 0.443 & 0.427 & 0.433 & 0.460 & 0.415 & 0.402 & 0.453 \\
LLaVA-1.5 & 3 & 0.433 & 0.427 & 0.445 & 0.455 & 0.445 & 0.442 & 0.462 & 0.460 & 0.468 & 0.453 \\
LLaVA-1.5 & 4 & 0.440 & 0.430 & 0.428 & 0.448 & 0.448 & 0.445 & 0.433 & 0.450 & 0.445 & 0.453 \\
\addlinespace[2pt]
LLaVA-NeXT & 2 & 0.522 & 0.512 & 0.503 & 0.540 & 0.545 & 0.548 & 0.553 & 0.510 & 0.543 & 0.553 \\
LLaVA-NeXT & 3 & 0.553 & 0.532 & 0.565 & 0.558 & 0.538 & 0.542 & 0.558 & 0.550 & 0.553 & 0.553 \\
LLaVA-NeXT & 4 & 0.560 & 0.550 & 0.555 & 0.558 & 0.547 & 0.558 & 0.558 & 0.552 & 0.555 & 0.553 \\
\bottomrule
\end{tabular}
\end{table}

\clearpage
\begin{table}[!htbp]
\centering
\caption{\textbf{Caption fidelity, $1/12$ budget, MMBench calibration.} K+V protection with $K=\operatorname{round}(N_{\rm KV}/12)$. ROUGE-L F1 against BF16 output on $200$ examples, not human-caption quality. U = uniform; Out = output-only; HG = hybrid. Mean averages all eight models.}
\label{tab:budget_d12_mmbench_caption}
\setlength{\tabcolsep}{4.25pt}
\renewcommand{\arraystretch}{1.12}
\begin{tabular}{@{}lc*{10}{c}@{}}
\toprule
 & & \multicolumn{3}{c}{min--max} & \multicolumn{3}{c}{AKVQ-style} & \multicolumn{3}{c}{KIVI-style} & \\
\cmidrule(lr){3-5}\cmidrule(lr){6-8}\cmidrule(lr){9-11}
Model & Bits & U & Out & HG & U & Out & HG & U & Out & HG & BF16 \\
\midrule
LLaVA-1.5 & 2 & 0.373 & 0.476 & 0.435 & 0.646 & 0.632 & 0.653 & 0.638 & 0.641 & 0.652 & 1.000 \\
LLaVA-1.5 & 3 & 0.738 & 0.735 & 0.757 & 0.788 & 0.796 & 0.809 & 0.791 & 0.791 & 0.795 & 1.000 \\
LLaVA-1.5 & 4 & 0.847 & 0.827 & 0.831 & 0.891 & 0.883 & 0.893 & 0.873 & 0.885 & 0.873 & 1.000 \\
\addlinespace[2pt]
LLaVA-NeXT & 2 & 0.460 & 0.475 & 0.492 & 0.626 & 0.622 & 0.653 & 0.551 & 0.553 & 0.584 & 1.000 \\
LLaVA-NeXT & 3 & 0.724 & 0.746 & 0.765 & 0.825 & 0.843 & 0.845 & 0.813 & 0.805 & 0.821 & 1.000 \\
LLaVA-NeXT & 4 & 0.885 & 0.871 & 0.882 & 0.902 & 0.888 & 0.898 & 0.881 & 0.873 & 0.881 & 1.000 \\
\bottomrule
\end{tabular}
\end{table}

\clearpage
\begin{table}[!htbp]
\centering
\caption{\textbf{Six-task accuracy, $1/4$ budget, POPE calibration.} K+V protection with $K=\operatorname{round}(N_{\rm KV}/4)$. Equal-weight accuracy over POPE, MME, MMBench, ScienceQA, MMMU, and SEED; up to $1000$ questions/task ($900$ for POPE; $839$ for MMMU). U = uniform; Out = output-only; HG = hybrid. Mean averages all eight models.}
\label{tab:budget_d4_pope_six}
\setlength{\tabcolsep}{4.25pt}
\renewcommand{\arraystretch}{1.12}
\begin{tabular}{@{}lc*{10}{c}@{}}
\toprule
 & & \multicolumn{3}{c}{min--max} & \multicolumn{3}{c}{AKVQ-style} & \multicolumn{3}{c}{KIVI-style} & \\
\cmidrule(lr){3-5}\cmidrule(lr){6-8}\cmidrule(lr){9-11}
Model & Bits & U & Out & HG & U & Out & HG & U & Out & HG & BF16 \\
\midrule
LLaVA-1.5 & 2 & 0.515 & 0.579 & 0.611 & 0.637 & 0.637 & 0.639 & 0.638 & 0.637 & 0.642 & 0.643 \\
LLaVA-1.5 & 3 & 0.640 & 0.642 & 0.645 & 0.642 & 0.645 & 0.643 & 0.644 & 0.642 & 0.645 & 0.643 \\
LLaVA-1.5 & 4 & 0.643 & 0.645 & 0.642 & 0.644 & 0.645 & 0.643 & 0.642 & 0.643 & 0.642 & 0.643 \\
\addlinespace[2pt]
LLaVA-NeXT & 2 & 0.618 & 0.658 & 0.669 & 0.663 & 0.667 & 0.667 & 0.656 & 0.666 & 0.668 & 0.666 \\
LLaVA-NeXT & 3 & 0.667 & 0.669 & 0.670 & 0.668 & 0.665 & 0.666 & 0.666 & 0.666 & 0.668 & 0.666 \\
LLaVA-NeXT & 4 & 0.667 & 0.666 & 0.666 & 0.665 & 0.667 & 0.667 & 0.667 & 0.666 & 0.665 & 0.666 \\
\bottomrule
\end{tabular}
\end{table}

\clearpage
\begin{table}[!htbp]
\centering
\caption{\textbf{TextVQA, $1/4$ budget, POPE calibration.} K+V protection with $K=\operatorname{round}(N_{\rm KV}/4)$. Custom VQA-style score on $200$ questions. U = uniform; Out = output-only; HG = hybrid. Mean averages all eight models.}
\label{tab:budget_d4_pope_textvqa}
\setlength{\tabcolsep}{4.25pt}
\renewcommand{\arraystretch}{1.12}
\begin{tabular}{@{}lc*{10}{c}@{}}
\toprule
 & & \multicolumn{3}{c}{min--max} & \multicolumn{3}{c}{AKVQ-style} & \multicolumn{3}{c}{KIVI-style} & \\
\cmidrule(lr){3-5}\cmidrule(lr){6-8}\cmidrule(lr){9-11}
Model & Bits & U & Out & HG & U & Out & HG & U & Out & HG & BF16 \\
\midrule
LLaVA-1.5 & 2 & 0.303 & 0.380 & 0.363 & 0.443 & 0.442 & 0.450 & 0.460 & 0.420 & 0.395 & 0.453 \\
LLaVA-1.5 & 3 & 0.433 & 0.438 & 0.462 & 0.455 & 0.457 & 0.458 & 0.462 & 0.457 & 0.462 & 0.453 \\
LLaVA-1.5 & 4 & 0.440 & 0.453 & 0.438 & 0.448 & 0.453 & 0.443 & 0.433 & 0.442 & 0.447 & 0.453 \\
\addlinespace[2pt]
LLaVA-NeXT & 2 & 0.522 & 0.518 & 0.492 & 0.540 & 0.530 & 0.535 & 0.553 & 0.537 & 0.525 & 0.553 \\
LLaVA-NeXT & 3 & 0.553 & 0.557 & 0.542 & 0.558 & 0.555 & 0.552 & 0.558 & 0.567 & 0.563 & 0.553 \\
LLaVA-NeXT & 4 & 0.560 & 0.558 & 0.558 & 0.558 & 0.550 & 0.543 & 0.558 & 0.563 & 0.557 & 0.553 \\
\bottomrule
\end{tabular}
\end{table}

\clearpage
\begin{table}[!htbp]
\centering
\caption{\textbf{Caption fidelity, $1/4$ budget, POPE calibration.} K+V protection with $K=\operatorname{round}(N_{\rm KV}/4)$. ROUGE-L F1 against BF16 output on $200$ examples, not human-caption quality. U = uniform; Out = output-only; HG = hybrid. Mean averages all eight models.}
\label{tab:budget_d4_pope_caption}
\setlength{\tabcolsep}{4.25pt}
\renewcommand{\arraystretch}{1.12}
\begin{tabular}{@{}lc*{10}{c}@{}}
\toprule
 & & \multicolumn{3}{c}{min--max} & \multicolumn{3}{c}{AKVQ-style} & \multicolumn{3}{c}{KIVI-style} & \\
\cmidrule(lr){3-5}\cmidrule(lr){6-8}\cmidrule(lr){9-11}
Model & Bits & U & Out & HG & U & Out & HG & U & Out & HG & BF16 \\
\midrule
LLaVA-1.5 & 2 & 0.373 & 0.511 & 0.545 & 0.646 & 0.668 & 0.722 & 0.638 & 0.644 & 0.693 & 1.000 \\
LLaVA-1.5 & 3 & 0.738 & 0.766 & 0.776 & 0.788 & 0.822 & 0.829 & 0.791 & 0.810 & 0.812 & 1.000 \\
LLaVA-1.5 & 4 & 0.847 & 0.857 & 0.862 & 0.891 & 0.895 & 0.893 & 0.873 & 0.880 & 0.896 & 1.000 \\
\addlinespace[2pt]
LLaVA-NeXT & 2 & 0.460 & 0.500 & 0.524 & 0.626 & 0.675 & 0.688 & 0.551 & 0.606 & 0.606 & 1.000 \\
LLaVA-NeXT & 3 & 0.724 & 0.797 & 0.812 & 0.825 & 0.872 & 0.850 & 0.813 & 0.816 & 0.820 & 1.000 \\
LLaVA-NeXT & 4 & 0.885 & 0.880 & 0.883 & 0.902 & 0.900 & 0.910 & 0.881 & 0.875 & 0.881 & 1.000 \\
\bottomrule
\end{tabular}
\end{table}

\clearpage
\begin{table}[!htbp]
\centering
\caption{\textbf{Six-task accuracy, $1/4$ budget, MMBench calibration.} K+V protection with $K=\operatorname{round}(N_{\rm KV}/4)$. Equal-weight accuracy over POPE, MME, MMBench, ScienceQA, MMMU, and SEED; up to $1000$ questions/task ($900$ for MMBench; $839$ for MMMU). U = uniform; Out = output-only; HG = hybrid. Mean averages all eight models.}
\label{tab:budget_d4_mmbench_six}
\setlength{\tabcolsep}{4.25pt}
\renewcommand{\arraystretch}{1.12}
\begin{tabular}{@{}lc*{10}{c}@{}}
\toprule
 & & \multicolumn{3}{c}{min--max} & \multicolumn{3}{c}{AKVQ-style} & \multicolumn{3}{c}{KIVI-style} & \\
\cmidrule(lr){3-5}\cmidrule(lr){6-8}\cmidrule(lr){9-11}
Model & Bits & U & Out & HG & U & Out & HG & U & Out & HG & BF16 \\
\midrule
LLaVA-1.5 & 2 & 0.514 & 0.605 & 0.633 & 0.639 & 0.639 & 0.641 & 0.639 & 0.644 & 0.641 & 0.644 \\
LLaVA-1.5 & 3 & 0.641 & 0.643 & 0.643 & 0.643 & 0.646 & 0.642 & 0.645 & 0.645 & 0.646 & 0.644 \\
LLaVA-1.5 & 4 & 0.644 & 0.644 & 0.645 & 0.646 & 0.645 & 0.642 & 0.643 & 0.645 & 0.644 & 0.644 \\
\addlinespace[2pt]
LLaVA-NeXT & 2 & 0.620 & 0.645 & 0.668 & 0.663 & 0.662 & 0.668 & 0.656 & 0.661 & 0.666 & 0.666 \\
LLaVA-NeXT & 3 & 0.666 & 0.669 & 0.669 & 0.668 & 0.666 & 0.667 & 0.667 & 0.667 & 0.667 & 0.666 \\
LLaVA-NeXT & 4 & 0.668 & 0.666 & 0.668 & 0.666 & 0.666 & 0.668 & 0.668 & 0.667 & 0.666 & 0.666 \\
\bottomrule
\end{tabular}
\end{table}

\clearpage
\begin{table}[!htbp]
\centering
\caption{\textbf{TextVQA, $1/4$ budget, MMBench calibration.} K+V protection with $K=\operatorname{round}(N_{\rm KV}/4)$. Custom VQA-style score on $200$ questions. U = uniform; Out = output-only; HG = hybrid. Mean averages all eight models.}
\label{tab:budget_d4_mmbench_textvqa}
\setlength{\tabcolsep}{4.25pt}
\renewcommand{\arraystretch}{1.12}
\begin{tabular}{@{}lc*{10}{c}@{}}
\toprule
 & & \multicolumn{3}{c}{min--max} & \multicolumn{3}{c}{AKVQ-style} & \multicolumn{3}{c}{KIVI-style} & \\
\cmidrule(lr){3-5}\cmidrule(lr){6-8}\cmidrule(lr){9-11}
Model & Bits & U & Out & HG & U & Out & HG & U & Out & HG & BF16 \\
\midrule
LLaVA-1.5 & 2 & 0.303 & 0.405 & 0.370 & 0.443 & 0.445 & 0.440 & 0.460 & 0.420 & 0.430 & 0.453 \\
LLaVA-1.5 & 3 & 0.433 & 0.442 & 0.443 & 0.455 & 0.465 & 0.447 & 0.462 & 0.453 & 0.467 & 0.453 \\
LLaVA-1.5 & 4 & 0.440 & 0.445 & 0.435 & 0.448 & 0.453 & 0.438 & 0.433 & 0.437 & 0.445 & 0.453 \\
\addlinespace[2pt]
LLaVA-NeXT & 2 & 0.522 & 0.523 & 0.537 & 0.540 & 0.522 & 0.518 & 0.553 & 0.515 & 0.522 & 0.553 \\
LLaVA-NeXT & 3 & 0.553 & 0.533 & 0.555 & 0.558 & 0.540 & 0.547 & 0.558 & 0.563 & 0.550 & 0.553 \\
LLaVA-NeXT & 4 & 0.560 & 0.555 & 0.548 & 0.558 & 0.540 & 0.558 & 0.558 & 0.547 & 0.547 & 0.553 \\
\bottomrule
\end{tabular}
\end{table}

\clearpage
\begin{table}[!htbp]
\centering
\caption{\textbf{Caption fidelity, $1/4$ budget, MMBench calibration.} K+V protection with $K=\operatorname{round}(N_{\rm KV}/4)$. ROUGE-L F1 against BF16 output on $200$ examples, not human-caption quality. U = uniform; Out = output-only; HG = hybrid. Mean averages all eight models.}
\label{tab:budget_d4_mmbench_caption}
\setlength{\tabcolsep}{4.25pt}
\renewcommand{\arraystretch}{1.12}
\begin{tabular}{@{}lc*{10}{c}@{}}
\toprule
 & & \multicolumn{3}{c}{min--max} & \multicolumn{3}{c}{AKVQ-style} & \multicolumn{3}{c}{KIVI-style} & \\
\cmidrule(lr){3-5}\cmidrule(lr){6-8}\cmidrule(lr){9-11}
Model & Bits & U & Out & HG & U & Out & HG & U & Out & HG & BF16 \\
\midrule
LLaVA-1.5 & 2 & 0.373 & 0.525 & 0.573 & 0.646 & 0.702 & 0.692 & 0.638 & 0.678 & 0.709 & 1.000 \\
LLaVA-1.5 & 3 & 0.738 & 0.777 & 0.781 & 0.788 & 0.831 & 0.835 & 0.791 & 0.830 & 0.830 & 1.000 \\
LLaVA-1.5 & 4 & 0.847 & 0.858 & 0.854 & 0.891 & 0.886 & 0.910 & 0.873 & 0.873 & 0.885 & 1.000 \\
\addlinespace[2pt]
LLaVA-NeXT & 2 & 0.460 & 0.518 & 0.544 & 0.626 & 0.682 & 0.677 & 0.551 & 0.602 & 0.649 & 1.000 \\
LLaVA-NeXT & 3 & 0.724 & 0.790 & 0.784 & 0.825 & 0.850 & 0.853 & 0.813 & 0.797 & 0.825 & 1.000 \\
LLaVA-NeXT & 4 & 0.885 & 0.887 & 0.882 & 0.902 & 0.900 & 0.892 & 0.881 & 0.864 & 0.873 & 1.000 \\
\bottomrule
\end{tabular}
\end{table}

\end{document}